\documentclass[journal]{IEEEtran}
\usepackage{amsmath,amsfonts}
\usepackage{algorithmic}
\usepackage{algorithm}
\usepackage{array}
\usepackage[caption=false,font=normalsize,labelfont=sf,textfont=sf]{subfig}
\usepackage{textcomp}
\usepackage{stfloats}
\usepackage{url}
\usepackage{hyperref}
\usepackage{verbatim}
\usepackage{graphicx}
\usepackage{cite}
\usepackage{xcolor}
\usepackage{makecell}
\usepackage{bm}
\usepackage{color}
\usepackage{multirow}
\definecolor{gray}{gray}{0.5}
\usepackage{amssymb}
\usepackage{amstext}

\usepackage[switch]{lineno}  %

\begin{document}

\title{Intra-class Adaptive Augmentation with Neighbor Correction for Deep Metric Learning}

\author{
Zheren Fu, 
Zhendong Mao,
Bo Hu,
An-An Liu, \IEEEmembership{Senior Member, IEEE}, \\
and Yongdong Zhang, \IEEEmembership{Senior Member, IEEE}

\thanks{
This work is supported in part by the National Natural Science Foundation of China (62222212, U19A2057, 62121002). 
Zhendong Mao is the corresponding author.
Zheren Fu is with the School of Cyberspace Science and Technology, University of Science and Technology of China, Hefei 230027, China (e-mail: fzr@mail.ustc.edu.cn).
Bo Hu is with the School of Information Science and Technology, University of Science and Technology of China, Hefei 230022, China (e-mail: hubo@ustc.edu.cn).  
Zhendong Mao and Yongdong Zhang are with the School of Information Science and Technology, University of Science and Technology of China and the Institute of Artificial Intelligence, Hefei Comprehensive National Science Center, Hefei 230022, China (e-mail: zdmao@ustc.edu.cn; zhyd73@ustc.edu.cn). 
An-An Liu is with School of Electrical and Information Engineering, Tianjin University, Tianjin 300072, China (e-mail: anan0422@gmail.com).}
}



\maketitle

\begin{abstract}
Deep metric learning aims to learn an embedding space, where semantically similar samples are close together and dissimilar ones are repelled against. 
To explore more hard and informative training signals for augmentation and generalization, recent methods focus on generating synthetic samples to boost metric learning losses.
However, these methods just use the deterministic and class-independent generations (e.g., simple linear interpolation), 
which only can cover the limited part of distribution spaces around original samples.
They have overlooked the wide characteristic changes of different classes and can not model abundant intra-class variations for generations.
Therefore, generated samples not only lack rich semantics within the certain class, but also might be noisy signals to disturb training.
In this paper, we propose a novel intra-class adaptive augmentation (IAA) framework for deep metric learning.
We reasonably estimate intra-class variations for every class and generate adaptive synthetic samples to support hard samples mining and boost metric learning losses.
Further, for most datasets that have a few samples within the class, we propose the neighbor correction to revise the inaccurate estimations, according to our correlation discovery where similar classes generally have similar variation distributions.
Extensive experiments on five benchmarks show our method significantly improves and outperforms the state-of-the-art methods on retrieval performances by 3\%-6\%.
Our code is available at 
\textcolor{blue}{\url{https://github.com/darkpromise98/IAA}}.
\end{abstract}


\section{Introduction}

\begin{figure}[t]
\centering
\includegraphics[width=1. \columnwidth]{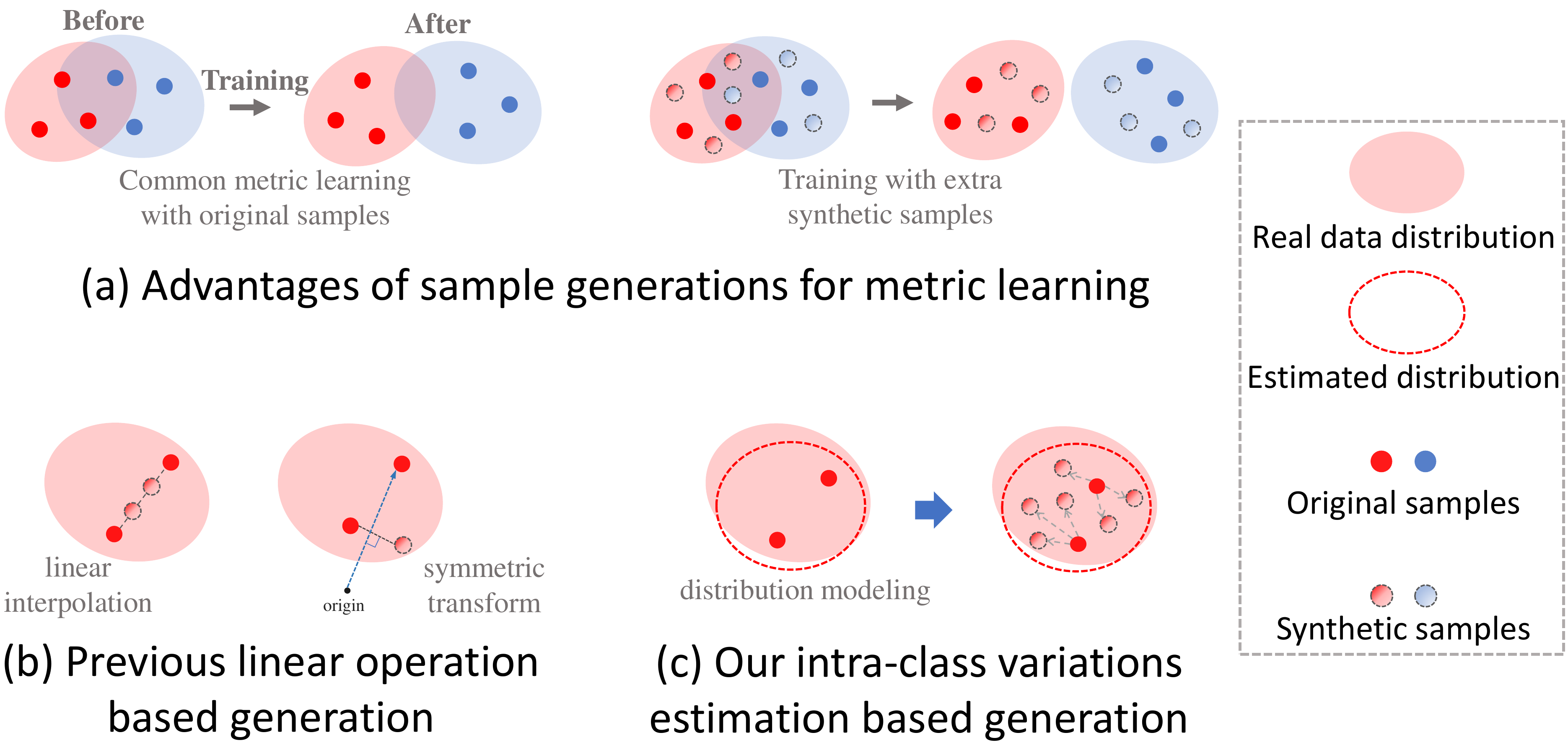}  
\caption{
\textbf{Illustration of sample generation, previous linear operation based generation methods, and our intra-class variations estimation based method for deep metric learning.}
(a) Sample generation based methods can produce and explore additional hard synthetic samples to learn a more discriminative embedding space \cite{duan2018deep}.
(b) Previous works \cite{gu2020symmetrical,ko2020embedding,gu2021proxy} use the deterministic linear operation to generate simple synthetic samples, which can not represent the diverse variations for each class.
(c) Our work properly considers the wide characteristic changes of different classes, and reasonably models the intra-class variation for generations.
Therefore, our generated samples have abundant and rational semantic property.
}
\label{motivation}
\end{figure}

Deep metric learning aims to learn an appropriate distance metric among arbitrary data and project them into an embedding space, where the embeddings of semantically similar samples are close together, while dissimilar ones are far apart. It is a crucial topic in multimedia areas and has been applied to various tasks, including image retrieval \cite{6197723,9113756,9507352}, person re-identification \cite{8506428,8985292}, detection \cite{8848601}, zero-shot learning \cite{7972969,9618856}, face recognition \cite{6151163,9479695}, cross-modal matching \cite{9320535,9178501}, and hashing \cite{hash1,hash2}.  
The main paradigm of deep metric learning is designing effective loss functions, which consider pair-wise relationships between different samples in the embedding space, 
such as Contrastive loss \cite{chopra2005learning}, Triplet loss \cite{schroff2015facenet},
and other elaborate losses \cite{wang2019multi,sun2020circle}. 
Metric learning losses get great improvements from various way, including sample mining  \cite{wu2017sampling,suh2019stochastic}, ensemble models \cite{milbich2020diva,rolinek2020optimizing}, and regularization techniques \cite{fu_2021_AAAI,chen2021graph}.
Besides, sample generation methods \cite{zhao2018adversarial,zheng2019hardness} have been proposed to alleviate models focus on a small minority of selected samples, and ignore a large majority of non-selected samples.
Various generated samples can provide extra informative training signals to support models learning a more discriminative embedding space, as shown in Fig. \ref{motivation}(a).
Existing works \cite{gu2020symmetrical,ko2020embedding,gu2021proxy} apply the algebraic computation in embedding spaces to generate samples and get powerful results.
In general, the essential challenge of generation-based methods is how to synthesize semantically rich samples with high confident labels.

\begin{figure*}[t]
\centering
\includegraphics[width=1. \textwidth]{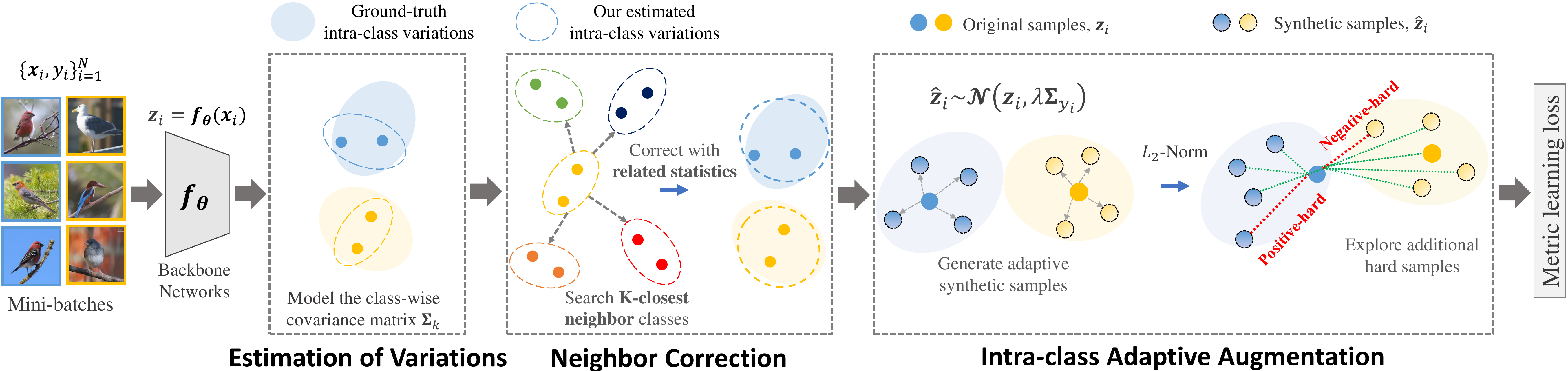}
\caption{
\textbf{Overview of our intra-class adaptive augmentation with neighbor correction framework.}
We first tentatively estimate the intra-class variations for each class in the embedding space, and use the neighbor correction to revise the inaccurate estimation for the few-samples class by integrating neighbor statistical information. 
Then we generate semantically diverse and meaningful synthetic samples and combine them with the sample mining to explore extra hard training signals.
Finally, we boost previous metric learning losses to get more discriminative embeddings and achieve better generalization.
}
\label{framework}
\end{figure*}


However, previous methods just use the deterministic and class-independent generations, 
e.g., simple linear interpolation and symmetric transform in Fig. \ref{motivation}(b),
which only can cover a tiny and limited part of distribution spaces around original samples.
They overlook the wide range of characteristic changes of different classes, and fail to model the diverse intra-class variations, which is significant for effective generations.
As a result, generated samples not only lack semantic diversity for the certain class, but also might be noisy signals during training,
which will hamper models to learn discriminative embeddings and generalization ability.
Besides, most datasets in metric learning and real scenes DO NOT have sufficient number of samples within the class (called few-samples class), thus the accurate estimation of intra-class variations is difficult by common statistical modeling. 
That's why existing generation-based methods just use deterministic and linear generation, which is not constrained by the number of samples within class, but have various weaknesses simultaneously. 




%

In this paper, we propose a novel intra-class adaptive augmentation (IAA) framework for deep metric learning, which is the first work to model rational intra-class variations for each class and generate semantically diverse and meaningful synthetic samples.
Different from previous deterministic generations, we present an efficient class-wise distribution estimation, which can be used to generate a variety of adaptive samples in Fig. \ref{motivation}(c). 
Further, for the few-samples class, we present the neighbor correction to solve the well-known challenge and get more accurate estimation.
Specifically, as shown in Fig. \ref{framework}, we first estimate the intra-class variations statistically, which can be regarded as the specific distributions for every class in the embedding space.
Then we use the neighbor correction to integrates neighbor statistical information and revise the defective estimations for the few-samples class,
in the light of our correlation discovery where similar classes generally have similar variation distributions.
Finally, we make use of estimated variations to generate adaptive samples and combine them with hard sample mining to boost metric learning losses.
Our method can be applied to any existing methods in a plug-and-play manner without any extra network modules, available for all kinds of datasets, and achieve significant performance improvements.
The contributions of this paper are four-fold:
\begin{itemize}
    
    \item We propose an efficient intra-class variations estimation approach for any form of datasets, which can properly model the class-wise distributions and support to generate adaptive hard samples for deep metric learning.
    
    \item We discover the strong correlation between classes and their variations, then present the neighbor correction to revise biased estimations for the few-samples class.
    
    \item  We propose the intra-class adaptive augmentation to generate semantically diverse and meaningful synthetic samples, then combine them with sample mining strategies to explore extra hard and informative representations.
    
    \item Extensive experiments on five typical benchmarks show our method greatly improves and outperforms the state-of-the-art approaches on retrieval tasks by 3\%-6\%.
\end{itemize}


\section{Related Work}
\label{related_wrok}

\subsection{Deep Metric Learning}

Deep Metric learning aims to learn an explicit non-linear mapping from raw data to a low dimensional embedding space, where similar (positive) samples are pushed together and dissimilar (negative) samples are repelled against, by the metric of the Euclidean distance. 
Plenty of loss functions have been proposed to get robust embeddings, and can be distinguished into categories, pair-based and proxy-based. 

Pair-based losses leverage the pairs of samples to build constraints.
Contrastive loss \cite{chopra2005learning} is the most typical method, which minimizes distances between positive pairs and maximizes distances between negative pairs.
In parallel, Triplet loss \cite{schroff2015facenet} considers three-tuples samples, and requires the distance of a negative pair to be higher than a positive pair by a fixed margin. 
Then N-pair \cite{sohn2016improved} and Lifted-structured loss \cite{oh2016deep} are proposed to exploit multiple samples and richer pair-wise relations, have shown to reach great results.

Proxy-based losses \cite{zhai2018classification,qian2019softtriple} use the learnable proxy embeddings as the class representation and the part of network parameters.
They leverage data-to-proxy relations \cite{proxyanchor,movshovitz2017no}, which encourage each sample to be close to the proxies of the same class and far away from  different classes, instead of other samples.  
Since they optimize the global structure of the embedding space, have less computations and converge faster if the number of classes is minor, otherwise the parameters are unaccepted and may cause out of memory \cite{wang2020cross,proxyanchor}.

Besides, computaions of metric learning is infeasible in practical for large datasets, and the majority of samples are not informative, more works started to focus on the sample minings.
Conclusive evidences \cite{wu2017sampling} show that the design of smart sampling strategies is as important as the design of efficient loss functions.
Recent improved losses \cite{wang2019multi,wang2019ranked} are combined with various samples mining strategies \cite{wu2017sampling,suh2019stochastic}, which select hard and informative sample pairs and give them higher loss weighting.
But samples minings usually focus on the selected minority and overlook the non-selected majority, which can lead to a overfitting model \cite{gu2020symmetrical,ko2020embedding}.

\subsection{Sample Generation}
Recently, sample generation methods \cite{lin2018deep,zhao2018adversarial,Fu2022SelfSupervisedSR} are proposed to improve the loss functions and mining strategies of deep metric learning. 
They can provide various more informative samples and potential training signals \cite{duan2018deep,gu2020symmetrical}, which do not exist in the original data. 
The essential challenge of generation-based methods is how to synthesize semantically rich samples with high confident labels.
Early methods \cite{duan2018deep,zheng2019hardness} leverage generative adversarial networks or auto-encoders to produce synthetic samples, but lead to extra-large models and unstable training.
Recent works \cite{gu2020symmetrical,ko2020embedding,gu2021proxy} generate virtual samples or classes in embedding spaces by geometric relations.
They just use the simple algebraic computation, such as linear interpolation and symmetric transform, but ignore the broad characteristic changes in different classes. 
Further, we solve these problems by modeling the intra-class variation and generate adaptive samples without any complex networks.

\subsection{Semantic Augmentation}
Semantic augmentation is a common technique to regularize deep networks in various visual tasks, and has been widely used to improve feature generalization and reduce training overfitting. 
Previous works  \cite{bengio2013better} discover inherent variations in feature spaces along with certain directions essentially correspond to implementing meaningful semantic changes in the original data domain.
For instance, deep feature interpolation \cite{upchurch2017deep} uses linear interpolations of features from a pre-trained networks to edit the semantics of images.
ISDA \cite{wang2019idsa} estimates the class-distribution in embedding spaces, and leverages specific feature transform to achieve semantic data augmentation effectively for downstream tasks.
Therefore, estimating suitable distributions of deep features and augmenting features from the estimation can improve the generalization in many areas  \cite{cai2020jcl,yang2021free}.
However, these methods generally get poor performances on the scarce data in minority classes \cite{li2021MetaSAug}.
We make use of the basic principle of semantic augmentation, and further propose a novel distribution correction method for the few-samples class and long-tail datasets.

\section{Preliminaries}
\label{preliminary}
This section introduces the basic mathematical formulation of deep metric learning.
Let $\bm{\mathcal{X}}=\{\bm{x}_i\}_{i=1}^N$ denotes a train dataset in the original data space,
and $\mathcal{Y}=\{y_i\}_{i=1}^N \in[1,2, \ldots, C]$ are the corresponding class labels. 
Deep metric learning aims to learn a mapping $ \bm{\mathcal{X}} \to \bm{\mathcal{Z}}$, from the original data space $\bm{x}_i\ \in \bm{\mathcal{X}}$ to the $D$-dimensional embedding space $\bm{z}_i\ \in \bm{\mathcal{Z}}$ by deep neural networks $f_{\theta}$: $ f_{\theta}(\bm{x}_i) = \bm{z}_i$,
so that embeddings of similar data are close together and dissimilar ones are far apart.
Distance metric between two samples $\bm{z}_i, \bm{z}_j$ in the embedding space can be defined as:

\begin{equation}
\label{distance}
    d_{ij}=d(\bm{z}_i, \bm{z}_j)=\left\|\bm{z}_i-\bm{z}_j\right\|_{2},
\end{equation}
where $d(,)$ is the Euclidean distance between two embeddings. 
We also define the similarity metric by computing cosine similarity $s(\bm{z}_i, \bm{z}_j)= \bm{z}_i^T \bm{z}_j / ({\|\bm{z}_i\|}_{2}{\|\bm{z}_j\|}_{2}) $.
$L_2$-norm usually is used to embeddings, so two metrics are equivalent.  
Various metric learning losses have been proposed in recent years.

\textbf{Contrastive loss} \cite{chopra2005learning} is the basic metric learning loss, which only considers two-tuple sample relations. 
It pulls positive pairs close together, and pushes negative pairs far away until given thresholds respectively: 
\begin{equation}
\label{contrastive}
    \mathcal{L}_{Cont} = \frac{1}{n} \sum_{i=1}^n ( \sum_{  y_i=y_j }  [d_{ij}-\beta_{p}]_{+} +  \sum_{ y_i\neq y_k } [\beta_{k} - d_{ik}]_{+} ),
\end{equation}
where $[\cdot]_+$ are the hinge function and $n$ is the mini-batch size. Every sample $\bm{z}_i$ in the mini-batch is regarded as an anchor point, it needs to match all relevant positive points $\bm{z}_j$ and negative points $\bm{z}_k$ to compute the loss.

%

\textbf{Triplet loss} \cite{schroff2015facenet} extends the contrastive formalism by considering three-tuple relationships. It pulls the anchor point $\bm{z}_i$ closer to the positive point $\bm{z}_j$ than to the negative point $\bm{z}_k$ ($y_i=y_j\neq y_k$) by a fixed margin $m$:
\begin{equation}
\label{triplet}
    \mathcal{L}_{Trip} = \frac{1}{n} \sum_{i=1}^n  \sum_{(j,k) \in \mathcal{T}_i } [d_{ij}-d_{ik}+ \alpha]_{+} ,
\end{equation}
where triplet sets $\mathcal{T}_i$ are constructed from the current mini-batches by various sample mining \cite{oh2016deep,wu2017sampling}, 
which usually select hard positive samples with relatively larger distances, and hard negative samples with smaller distances.

\textbf{MS loss} \cite{wang2019multi} is one of the latest work for metric learning.
it measures both self-similarity and relative similarities of sample pairs,
which can explore informative samples in a mini-batch and weight them by multiple similarities (MS):



\begin{equation}
\label{ms}
    \begin{aligned}
\mathcal{L}_{MS} &=\frac{1}{n} \sum_{i=1}^n \left\{\frac{1}{\alpha} \log \left[  1+\sum_{j \in \mathcal{P}_{i} } e^{\alpha\left(\lambda - s_{ij}\right)} \right]\right. \\
&\left.+\frac{1}{\beta} \log \left[1+\sum_{k \in \mathcal{N}_{i} } e^{\beta \left(s_{ik}-\lambda\right)}\right]\right\},
\end{aligned}
\end{equation}
where $\alpha$, $\beta$, $\lambda$ are the loss hyper-parameters and $s_{ij}$ is cosine similarity.  $\mathcal{P}_{i}$ and $\mathcal{N}_{i}$ are selected hard positive and negative samples set for the anchor $\bm{z}_i$ by its hard sample mining.

\begin{equation}
\label{ms_p}
    \mathcal{N}_i = \{ s_{ij} | s_{ij} > \min _{y_k=y_i} s_{ik}-\epsilon\}
\end{equation}

\begin{equation}
\label{ms_n}
    \mathcal{P}_i = \{ s_{ij} | s_{ij} < \max _{y_k \neq  y_i} s_{ik} + \epsilon\}
\end{equation}

\section{Proposed Method} 
\label{proposed_method}

In this section, we first introduce the estimation of intra-class variations and related challenge for the few-samples class.
Then, we describe our correlation discovery between classes and their variations, 
which provides the foundation for the neighbor correction.
Finally, we integrate our intra-class adaptive augmentation (IAA) into existing methods.

\subsection{Estimation of Intra-class Variations}
To generate adaptive synthetic samples for each class, we need to estimate the intra-class variations properly.
We assume all samples of each class belong to the specific Gaussian distribution in the embedding space. 
Therefore, we use the Maximum Likelihood principle \cite{akaike1998information} to approximately estimate the mean and covariance matrix of the class-wise distributions.
If the class $k$ ($k \in [1,..., C]$) have the sample indices set $\mathbb{S}_k$, which contains all samples within the class, and the number of samples is $n_k$, we have:

\begin{equation}
\label{mean}
\boldsymbol{\mu}_{k}=\frac{1}{n_k} \sum_{i\in \mathbb{S}_k} \boldsymbol{z}_{i},
\end{equation}

\begin{equation}
\label{cov}
\boldsymbol{\Sigma}_{k}=\frac{1}{n_{k}} \sum_{i \in \mathbb{S}_k}\left(\boldsymbol{z}_{i}-\boldsymbol{\mu}_{k}\right)\left(\boldsymbol{z}_{i}-\boldsymbol{\mu}_{k}\right)^{T},
\end{equation}
where the covariance matrix $\boldsymbol{\Sigma}_{k}$ reflects the characteristic variations within the class $k$, 
and can be easily computed by aggregating all embeddings from the whole dataset.


\begin{figure}[t]
\centering
\includegraphics[width=1. \columnwidth]{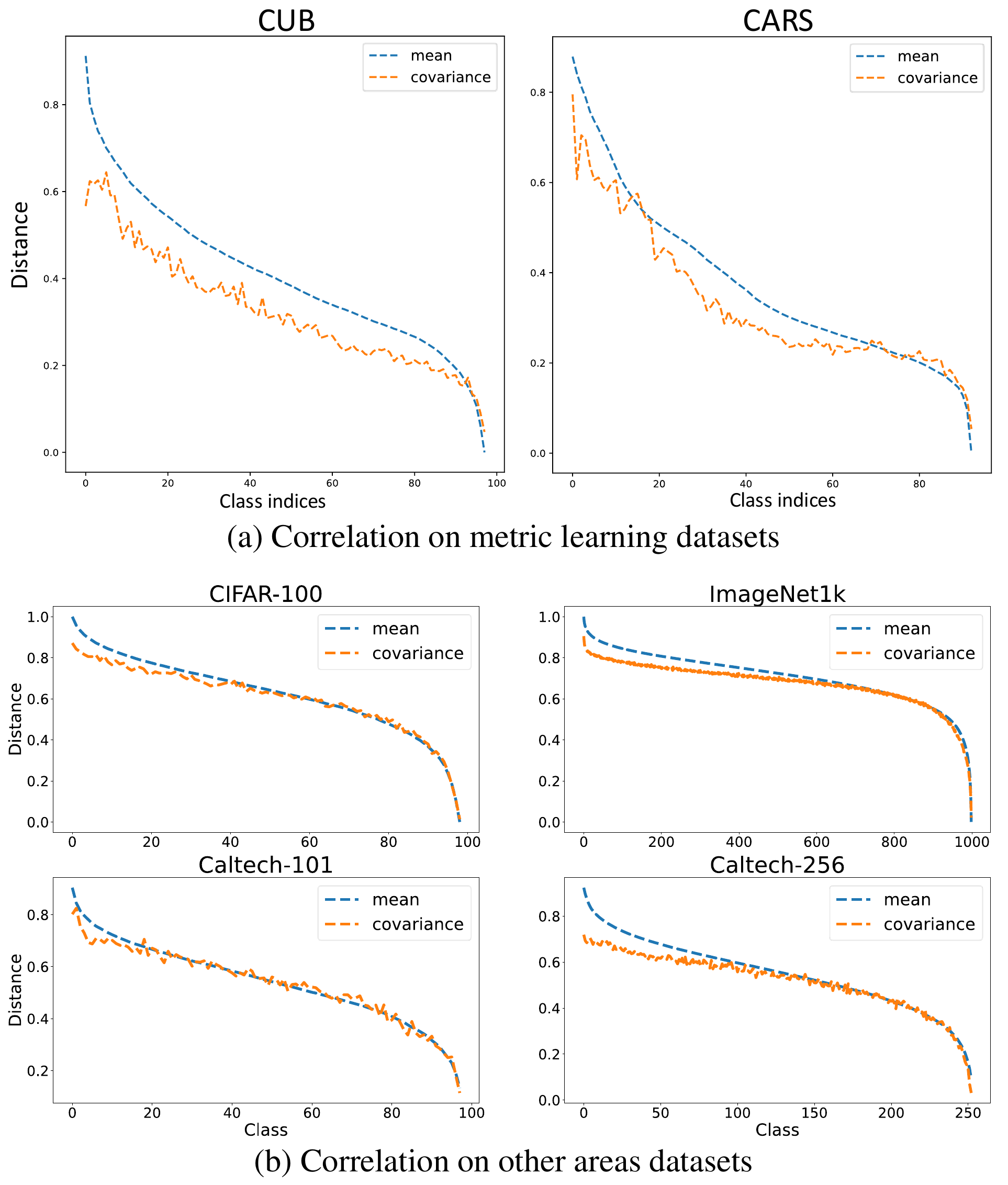}  
\caption{\textbf{Illustration of the strong correlation between mean distances and covariance distances.}  
We use the GoogLeNet \cite{szegedy2015going} pre-trained on ImageNet to extract their embeddings.
(a) Correction on two typical metric learning datasets, CUB \cite{WahCUB_200_2011} and CARS \cite{krause20133d}.
(b) Correlation on four other areas datasets, CIFAR-100 \cite{2012Learning}, ImageNet-1k \cite{russakovsky2015imagenet}, Caltech101\cite{caltech256}, and Caltech256\cite{caltech256}.
Class indices are re-listed according to the descending orders of mean distances (blue curve), 
and distances are normalized to $[0, 1]$.
We get the final curve by averaging the results on all classes.
For simplicity, we use the diagonal elements of covariance matrix to compute distances ($p$=2).
}
\label{ranking}
\end{figure}

However, the reasonable estimation of covariance matrices is not always easy. 
This common statistical method requires a sufficient amount of samples in every class for the accurate estimation \cite{wang2019idsa},
Unfortunately, only small datasets (e.g., CUB and CARS in metric learning) could have adequate number of samples in each class. 
Most of the large datasets (e.g., SOP, InShop, and VehicleID in metric learning) in real scenes only have scarce samples within the class, so the appropriate estimation becomes an ill-posed problem \cite{li2021MetaSAug}.

That's why latest synthetic sample generation methods \cite{gu2020symmetrical,ko2020embedding,gu2021proxy} just use simple linear operations to generate samples, it's not constrained by the number of samples within class, and have various limitations simultaneously.
To solve this problem, we first expound on our discovery of the strong statistical (positive) correlation from estimated distributions.

\subsection{Correlation Discovery}
\label{correlation}

Theoretically, the mean and covariance matrix of a certain Guassian distribution are independent parameters, 
since they are the two sufficient statistics \cite{lehmann2006theory}.
However, we have discovered their unique statistical correlation on our estimated class-wise distributions:
\textit{If the distances of different means are close, the distances of corresponding covariance matrices will be close, too.}
Formally, we use the $p$-norm of vector or matrix difference to define the distance metrics of mean and covariance, between the class $i$ and the class $j$:

\begin{equation}
\label{mean_dist}
    D_m(\bm{\mu}_i, \bm{\mu}_j)=\left\|\bm{\mu}_i^2 -\bm{\mu}_j^2\right\|_{p},
\end{equation}

\begin{equation}
\label{cov_dist}
    D_{cv}(\bm{\Sigma}_{i}, \bm{\Sigma}_{j})=\left\|\bm{\Sigma}_{i}-\bm{\Sigma}_{j}\right\|_{p},
\end{equation}

Given a certain class, we compute the mean distance and covariance distance with other classes.
Then we regard two types of distances as two variable sequences, and re-list the class indices  according to the descending order of mean distances. 
Fig. \ref{ranking} shows the related results on two typical metric learning datasets (CUB and CARS), and four datasets widely used in various visual tasks.
Besides, we compute the Spearman correlation coefficient \cite{2013Research}, which quantitatively evaluates the ranking correlation of two variables in Tab. \ref{tab:spearman}.

It is not hard to find that the mean distances and corresponding covariance matrices distances have the strong correlation \cite{yang2021free}.
That is to say, the similar means usually have similar covariance matrices in our estimated class-wise distributions.
Further, the similar classes usually have similar intra-class variations.
Therefore, we can use samples of similar classes to improve the inaccurate estimation.

\begin{table}[t]
\normalsize
\centering
\caption{
\textbf{Spearman correlation coefficient between mean distances and covariance distances.} 
The value is from 0 to 1, larger is better.
We show different distance metrics as Eq. (\ref{mean_dist}) and (\ref{cov_dist}).
}
\label{tab:spearman}
\begin{tabular}{l|cccc}
\hline
\multirow{2}{*}{Datasets} & \multicolumn{4}{c}{Spearman correlation coefficient}   \\
 & $p=1$ & $p=2$ & $p=3$ & $p=4$ \\ \hline  

CUB \cite{WahCUB_200_2011} & 0.81 & 0.77 & 0.74 & 0.69 \\
CARS  \cite{krause20133d} & 0.91 & 0.85 & 0.78 & 0.74  \\ \hline

CIFAR-100 \cite{2012Learning}               & 0.94 & 0.90 & 0.85 & 0.80 \\
ImageNet-1k \cite{russakovsky2015imagenet}  & 0.82 & 0.71 & 0.64 & 0.62  \\
Caltech101 \cite{caltech256}                & 0.93 & 0.88 & 0.80 & 0.72 \\
Caltech256 \cite{caltech256}                & 0.94 & 0.91 & 0.85 & 0.78  \\

\hline
\end{tabular}%
\end{table}

\subsection{Neighbor Correction}
\label{neighbor_correction}
For the few-samples class dataset, 
we use the k-nearest neighbor principle to assemble adequate statistical information from neighbor classes and correct the defective estimation,
based on our correlation discovery in Section  \ref{correlation}.

Without loss of generality, we use the diagonal covariance matrix and the $2$-norm as the distance metric for simple computation.
Given the class $k$, we select the top-$K$ most similar classes by computing the mean distance as Eq. (\ref{mean_dist}).

\begin{equation}
\mathbb{N}_{k}=\left\{i \mid   \left\|\bm{\mu}_i^2 -\bm{\mu}_k^2\right\|_{2} \in \operatorname{min(K)}, \ i \in [1,..., C] \right\},
\end{equation}
where $ \operatorname{min(K)}$ aims to select the minimal $K$ elements, 
so $\mathbb{N}_{k}$ represents the indices set of $K$ closest neighbor classes for the class $k$.
Then we compute the weighted sum by integrating all statistical information of neighbor classes:

\begin{equation}
\bm{\Sigma}_{k, \text { neighbor }}= (\sum_{i \in \mathbb{N}_{k}} w_{i} \bm{\Sigma}_{i}) /  (\sum_{i \in \mathbb{N}_{k}} w_{i}),
\end{equation}

$w_i$ is the weight for each $\bm{\Sigma}_{i}$,
and it is determined by three parts, the number of samples $n_k$, the relevant mean distance $D_m(\bm{\mu}_i, \bm{\mu}_k)$, and covariance distance $D_{cv}(\bm{\Sigma}_{i}, \bm{\Sigma}_{k})$:

\begin{equation}
\label{wi}
 w_{i} = n_i \cdot \exp{ (-\frac{\left\|\bm{\mu}_i^2 -\bm{\mu}_k^2\right\|_{2}^2}{2 \sigma_m^2} -\frac{\left\|\bm{\Sigma}_{i}-\bm{\Sigma}_{j}\right\|_{2}^2}{2 \sigma_{cv}^2}) },
\end{equation}

The number of samples $n_i$ can represent the proportion of neighbor statistics, 
and two relative distances of mean and covariance are constructed by the Gaussian kernel \cite{2002Bilateral}, 
where more similar neighbor classes will get larger weight coefficients.
$\sigma_{m}$ and $\sigma_{cv}$ control the spread of related distances.
If $\sigma_{m}=\sigma_{cv}=\infty$, Eq. (\ref{wi}) will degrade to the simple average without considering the distances of means and covariances.
Besides, we compute the global covariance matrix from all classes and use it to avoid overfitting. 

\begin{equation}
\bm{\Sigma}_{ \text { global }} = ( \sum_{k=1}^{C} n_{k} \bm{\Sigma}_{k} ) /( \sum_{k=1}^{C} n_{k} ),
\end{equation}

Finally, we introduce the neighbor correction to the initial covariance matrix for revising the defective estimation.

\begin{equation}
\label{revise}
\bm{\Sigma}_{k}^{cr} =(1-\alpha) \bm{\Sigma}_{k}+\alpha [ (1-\gamma) \bm{\Sigma}_{k, \text { neighbor }}  +  \gamma \bm{\Sigma}_{ \text { global }} ], 
\end{equation}
where $\alpha$ is a dynamic coefficient relied on the number of samples  $n_k$, and computed by the neighbor weight function: 

\begin{equation}
\label{nwf}
\alpha=\left\{\begin{array}{l}
(1+\log [1+\beta \cdot (n_k-1)])^{-1} , \ \ n_{k} \leq \tau, \\
0, \ \ n_{k} > \tau.
\end{array}\right.
\end{equation}

The larger number of samples within the class is, the smaller value $\alpha$ is, the lower effect of the neighbor correction is applied to the initial covariance matrix.
Therefore If $n_k$ is large, $\alpha$ will close to 0, the neighbor correlation is unnecessary.
$\gamma$ is a fixed coefficient to balance global and neighbor information.

\subsection{Intra-class Adaptive Augmentation}
\label{iaa}
With proper estimation and correction of intra-class variations as Eq. (\ref{cov}) and Eq. (\ref{revise}), we can generate adaptive synthetic samples.
Since the class-wise covariance matrix proportionally represents the intra-class variations, and contains rich semantic diversity \cite{em2017incorporating,wang2019idsa}. 
For each original sample $\bm{z}_i$, we generate synthetic samples with semantically meaningful changes, through translating $\bm{z}_i$ along different directions sampled from $\mathcal{N}\left(0, \lambda \Sigma_{y_{i}}\right)$ in the embedding space. Equivalently, we get:

\begin{equation}
\label{generation}
\hat{\boldsymbol{z}}_{i} \sim \mathcal{N}\left(\boldsymbol{z}_{i}, \lambda \boldsymbol{\Sigma}_{y_{i}}\right),
\end{equation}
where $\lambda$ is a positive coefficient to control the strength of intra-class variations. 
We record the multiple synthetic samples generated from original $\bm{z}_i$ as $\hat{\bm{z}}_i^m, m=1,..., M$. 
Then we combine with hard sample mining to explore more informative samples.
We still make original samples as the anchor, but regard all synthetic samples as additional positive and negative candidates for each anchor.
We formulate the boosting versions of metric learning losses in Section \ref{preliminary}.

\textbf{Contrastive + IAA} makes use of all sample-pairs in mini-batches.
The original positive samples set for anchor $\bm{z}_i$ is $\mathcal{P}_i$, and extra synthetic positive samples set is $\hat{\mathcal{P}}_i$. 
Similarly, $\mathcal{N}_i / \hat{\mathcal{N}}_i$ is the original/synthetic negative samples set. 
\begin{equation}
\label{contrastive2}
   \mathcal{L}_{Cont}^{IAA} = \frac{1}{n} \sum_{i=1}^n ( \sum_{  j \in \mathcal{P}_i \cup \hat{\mathcal{P}}_i }  [d_{ij}-\beta_{p}]_{+} +  \sum_{k \in \mathcal{N}_i \cup \hat{\mathcal{N}}_i} [\beta_{k} - d_{ik}]_{+} ),
\end{equation}

\textbf{Triplet + IAA} is combined with the sample mining, we add synthetic samples as extra candidates to select the hardest negative and all available positive samples.
So synthetic samples can provide more three-tuple and informative training signals. 
For each anchor $\bm{z}_i$, we mark $\hat{\mathcal{T}}_i$ as the triplet set from synthetic positive samples  $\hat{\bm{z}}_p^m$ and negative samples $\hat{\bm{z}}_n^m$.
\begin{equation}
\label{triplet2}
    \mathcal{L}_{Trip}^{IAA} =   \frac{1}{n} \sum_{i=1}^n ( \sum_{(j,k) \in  \mathcal{T}_i  \cup \hat{\mathcal{T}}_i  } [ {(d_{ij})}-\min_k {(d_{ik})}+ \alpha]_{+}),
\end{equation}

\textbf{MS + IAA} also extends more informative sample-pairs by synthetic samples. For each anchor $\bm{z}_i$, we mark extra synthetic set of positive samples as $\hat{\mathcal{P}}_i$ and negative as $\hat{\mathcal{N}}_i$ according to Eq. (\ref{ms_p}) and Eq. (\ref{ms_n}), then add them to construct losses.

\begin{equation}
\label{ms2}
    \begin{aligned}
\mathcal{L}_{MS}^{IAA} &=\frac{1}{n} \sum_{i=1}^n \left\{\frac{1}{\alpha} \log \left[  1+\sum_{j \in \mathcal{P}_{i} \cup \hat{\mathcal{P}}_i } e^{\alpha\left(\lambda - s_{ij}\right)} \right]\right. \\
&\left.+\frac{1}{\beta} \log \left[1+\sum_{k \in \mathcal{N}_{i}  \cup \hat{\mathcal{N}}_i } e^{\beta \left(s_{ik}-\lambda\right)}\right]\right\},
\end{aligned}
\end{equation}

Finnaly, the whole training process of our proposed method is outlined in Algorithm \ref{alg}.

\begin{algorithm}[t]
    \caption{Train a model with our framework}
    \label{alg}
        \begin{algorithmic} 
        \REQUIRE  images $\bm{\mathcal{X}}$, class labels $\bm{\mathcal{Y}}$, network model $f_{\theta}$ \\ 
        \WHILE{Not Converged} 
                \STATE Estimate the covariance matrix $\bm{\Sigma}_{k}$ as Eq. (\ref{cov}).
               \IF{Belong the few-samples class}
                    \STATE Apply the neighbor correction as Eq. (\ref{revise}).
                \ENDIF 
                \FOR{$t=0$ to $T$}
                    \STATE Construct a mini-batch $ \{\bm{x}_i, y_i\}_{i=1}^n$ from $\bm{\mathcal{X}}, \bm{\mathcal{Y}}$
                    \STATE Compute embeddings $ \bm{z}_i = f_{\theta}(\bm{x}_i)$
                    \STATE Generate adaptive samples $\hat{\bm{z}}_i$ as Eq. (\ref{generation})
                    \STATE Boost metric learning losses as Eq. (\ref{contrastive2}) or Eq. (\ref{triplet2})  
                    \STATE Optimize model parameters $\theta$. 
                \ENDFOR
        \ENDWHILE
        
        \end{algorithmic}
    \end{algorithm}

\section{Analysis}

\subsection{Comparison to Existing Works}
Our work belongs to the generation-based metric learning methods, which produce the synthetic samples in the embedding spaces.
However, both our core motivation and technical implement fundamentally differ from previous methods \cite{lin2018deep,zhao2018adversarial,duan2018deep,ko2020embedding,gu2020symmetrical,gu2021proxy}.
Existing methods just use the deterministic and class-independent generations, 
(e.g., simple linear interpolation and symmetric transform)
which only can cover a tiny and limited part of distribution spaces around original samples.
In other words, they overlook the wide range of characteristic changes of different classes, and fail to model the diverse intra-class variations, which is significant for effective generations.
In contrast, our method not only can generate samples with rich semantic diversity for the certain class, but also alleviate the noisy signals during training.
Therefore, our models can learn more discriminative embeddings and generalization ability, compared with previous methods.

Besides, our method is close to semantic augmentation, which performs semantic changes  for enhancing  generalization and reduce overfitting.
Previous works  \cite{upchurch2017deep,wang2019idsa,cai2020jcl,yang2021free} implement specific feature transform or estimate feature distributions in embedding spaces to achieve semantic augmentation. 
But these methods do not consider the scarce data in minority classes, and can not get reasonable results on the few-samples class.
In comparison, we propose a novel distribution correction method for the few-samples class and long-tail datasets to solve this problem.

\subsection{Ground of correlation}
Intuitively, our correlation discovery is based on this ground: the deep feature learning of visual tasks aims to ensure that the distance of image features can reflect the similarity of image semantics, i.e., image features are the numerical representation of image semantics. 
So the changes of image semantics can be reflected on the variance of image features. 
In real scene, these semantical changes in similar classes are similar, 
e.g., the posture changes between different cats are similar, compared to other animals. 
Therefore, features of similar classes have similar variance distribution.

Besides, we find on the few-shot learning \cite{yang2021free,yang2021free2} and long-tail classification  \cite{Liu2020DeepRL,Hu2021LeveragingTF} fields,
the similar phenomenon is observed, that is the similar classes usually have similar mean and variance of features.
These works can provide sufficient guidances and supports for our correlation discovery,
which is rigorous enough for metric learning.

More profound analysis and experiments is on Section \ref{analysis}.
We provide more correlation results on different datasets (Section \ref{more_dataset} and \ref{distance_metric}), and show how the correlation affects the embedding learning and final performances (Section \ref{covariance_matrix} and \ref{available_nc}).
These experiments can prove the authority and indispensability of correction discovery.

\begin{table}[t]
    \centering
    \caption{
\textbf{The statistical information of our used datasets in train set.}
The first two are small datasets and have plenty of positive samples in each class. 
The last three are large datasets and have a few samples in each class. 
'max', 'min', 'avg', and 'std' means maximum, minimum, mean, and standard deviation respectively.
    }   
    \resizebox{1. \columnwidth}{!}{
\begin{tabular}{l|c|cc|cccc} 
    \hline
        &   &   &   & \multicolumn{4}{c}{Number of samples in class} \\
    Datasets     & Objects   & Samples &  Classes  & max  & min  & avg  & std \\  \hline  \hline 
    CUB  & birds   & 5,864   & 100 & 60 & 41 & \textbf{58.6} & 3.5  \\ 				
    CARS & cars   & 8,054    & 98 &  97  & 59 & \textbf{82.2} & 7.2  \\ \hline 			
    SOP  & furnitures & 59,551  & 11,318  & 12 & 2  & \textbf{5.3}  & 3.0 \\ 
    InShop & clothes & 25,882    & 3,997 & 162  & 1 & \textbf{6.5}  & 6.4 \\ 
    VehicleID & vehicles & 110,178    & 13,134 & 142 & 2 & \textbf{8.4}  & 9.5  \\ \hline 
\end{tabular}
    }
    \label{tab:dataset-info}
    \end{table}

\begin{table*}[t]
    \normalsize
    \centering
    \caption{
    \textbf{Retrieval performances of our method on five datasets and three baseline losses.}
    We list the previous sample generation methods, and all of them (including ours) use the GoogLeNet backbone with 512 embedding sizes for a fair comparison.
    $ReImp$ indicates our re-implementation with the standard settings \cite{wang2020cross}. 
    We use the Medium size of test set for PKU VehicleID.
    }
    \label{tab:baslines}
    \resizebox{1.0 \textwidth}{!}{
    \begin{tabular}{l|ccc|ccc|ccc|ccc|cc}
    \hline
      & \multicolumn{3}{c|}{CUB-200-2011} & \multicolumn{3}{c|}{Cars-196}  & \multicolumn{3}{c|}{Stanford Online Products} & \multicolumn{3}{c|}{In-shop Clothes Retrieval}  & \multicolumn{2}{c}{PKU VehicleID} \\
    Method        & R@1   & R@2  & R@4       & R@1  & R@2  & R@4      & R@1 & R@10 & R@100   & R@1 & R@10 & R@20  & R@1 & R@5      \\ \hline \hline
    
    Triplet+DAML \cite{duan2018deep}             & 37.6 & 49.3 & 61.3    & 60.6 & 72.5 & 82.5    & 58.1 & 75.0 & 88.0   & - & - & -  & - & -  \\
    Triplet+HDML \cite{zheng2019hardness}               & 43.6 & 55.8 & 67.7    & 61.0 & 72.6 & 80.7    & 58.5 & 75.5 & 88.3   & - & - & - & - & -    \\
    
    Triplet+DVML \cite{lin2018deep}                     & 43.7 & 56.0 & 67.8    & 64.3 & 73.7 & 79.2    & 66.5 & 82.3 & 91.8   & - & - & - & - & -  \\
    Triplet+Symm \cite{gu2020symmetrical}               & 55.0 & 67.3 & 77.5    & 69.7 & 78.7 & 86.1    & 68.5 & 82.4 & 91.3   & - & - & - & - & -  \\  
    
    Triplet+EE  \cite{ko2020embedding}                 & 51.7 & 63.5 & 74.5    & 71.6 & 80.7 & 87.5    & \textbf{77.2} & 89.6 & 95.5   & - & - & - & - & -   \\

    Triplet ($ReImp$)  & 54.4 & 65.8 & 76.0   & 69.0 & 78.1 & 85.4      & 73.5 & 87.8 & 94.9   & 86.1 & 95.7 & 97.0  & 83.5 & 93.3  \\
    \textbf{Triplet + IAA}          & \textbf{58.9}  & \textbf{70.6}  & \textbf{80.5}     & \textbf{72.4}  & \textbf{81.5}  & \textbf{88.2}        & 76.9  &  \textbf{89.9} & \textbf{96.0}   & \textbf{88.6}  & \textbf{97.4}  & \textbf{98.3} & \textbf{87.7} & \textbf{94.7}  \\ \hline
    
    Contrastive  ($ReImp$)          & 57.3  & 68.6  & 79.5     & 74.1 & 82.8 & 89.0         & 75.0  & 88.3  & 95.2   & 86.5  & 96.0  & 97.3 & 87.9 & 94.9  \\
    \textbf{Contrastive + IAA}     & \textbf{61.7}  & \textbf{73.3}  & \textbf{82.2}     & \textbf{80.8}  & \textbf{87.9}  & \textbf{92.8}        & \textbf{77.8}  &  \textbf{90.2} & \textbf{96.1}   & \textbf{88.8}  & \textbf{97.3}  & \textbf{98.2} & \textbf{91.6} & \textbf{95.9}   \\ \hline
    
    MS+EE \cite{gu2020symmetrical}               & 57.4 & 68.7 & 79.5    & 76.1 & 84.2 & 89.8    & 78.1 & 90.3 & 95.8   & - & - & -  & - & -  \\
    MS+PS  \cite{gu2021proxy}                 & 55.9 & 68.1 & 78.1    & 71.8 & 80.9 & 87.6    & - & - & -   & - & - & - & - & -   \\
    MS+L2A  \cite{Lee2022LearningTG}                 & 60.6 & 72.5 & 82.2    & 81.2 & 88.4 & 93.0    & - & - & -   & - & - & - & - & -   \\
    MS ($ReImp$)                 & 56.2 & 68.3 & 79.1   & 77.0 & 84.3 & 89.9    & 75.9 & 89.3 & 95.6   & 87.0 & 97.0 & 98.1 & 89.5 & 94.9   \\ 
    \textbf{MS + IAA}         & \textbf{62.1}  & \textbf{73.8}  & \textbf{82.7}     & \textbf{84.2}  & \textbf{90.4}  & \textbf{94.3}        & \textbf{78.6}  &  \textbf{91.0} & \textbf{96.5}   & \textbf{89.3}  & \textbf{97.8}  & \textbf{98.6} & \textbf{92.2} & \textbf{96.2}  \\ \hline  
    
    \end{tabular}
    }
    \end{table*}

\begin{table*}[t]
\normalsize
\centering
\caption{
\textbf{Comparison with the state-of-the-art methods on four datasets.}
Backbones are denoted by abbreviations: BN–InceptionBN, R–ResNet50. 
The superscripts represent embedding sizes. 
The best and second best results are marked in bold and underline.
}
\label{tab:sota1}
\resizebox{0.9 \textwidth}{!}{
\begin{tabular}{lc|ccc|ccc|ccc|ccc}
\hline
                   &     & \multicolumn{3}{c|}{CUB-200-2011} & \multicolumn{3}{c|}{Cars-196}   & \multicolumn{3}{c|}{Stanford Online Products} & \multicolumn{3}{c}{In-shop Clothes Retrieval}  \\
Method             &  Setting   & R@1 & R@2 & R@4 & R@1 & R@2 & R@4 & R@1 & R@10 & R@100 & R@1 & R@10 & R@20  \\ \hline \hline

Circle \cite{sun2020circle}      & BN$^{512}$    & 66.7 & 77.4 & 86.2 & 83.4 & 89.8 & 94.1 & 78.3 & 90.5 & 96.1 & - & - & - \\
XBM \cite{wang2020cross}         & BN$^{512}$   & 65.8 & 75.9 & 84.0  & 82.0 & 88.7 & 93.1  & 79.5 & 90.8 & 96.1    & 89.9 & 97.6 & 98.4  \\
PA \cite{proxyanchor}   & BN$^{512}$   & 68.4 & \underline{79.2} & \textbf{86.8}  & 86.1 & 91.7 & 95.0  & 79.1 & 90.8 & 96.2       & 91.5 & 98.1 & 98.8 \\
CGML \cite{chen2021graph}        & BN$^{512}$   & 66.3 & 77.2 & 86.3  & 85.6 & 91.1 & 94.2    & 79.1 & 90.8 & 96.1         & 90.5 & 98.2 & 98.6  \\
PS  \cite{gu2021proxy}           & BN$^{512}$   & 66.0 & 76.6 & 85.0 & 84.7 & 90.7 & 94.6   & 79.8 & 90.9 & 96.4         & 90.9 & 97.9 & 98.5  \\

FastAP \cite{cakir2019deep}                  & R$^{512}$    & - & - & - & - & - & -     & 76.4 & 89.0 & 95.1    & 90.9 & 97.7 & 98.5  \\
RaMBO \cite{rolinek2020optimizing}           & R$^{512}$    & 63.5 & 74.8 & 84.1 & - & - & -    & 78.6 & 90.5 & 96.0    & 88.1 & 97.0 & 97.9     \\
DiVA \cite{milbich2020diva}                  & R$^{512}$    & \textbf{69.2} & \textbf{79.3} & - & \underline{87.7} & \underline{92.9} & -    & 79.6 & 91.2 & -         & -    & -    & -     \\
SR \cite{fu_2021_AAAI}                       & R$^{512}$    & 68.2 & 78.1 & 86.5 & \underline{87.7} & 92.5 & \underline{95.4}     & 78.6 & 90.6 & 96.2    & -    & -    & -     \\ \hline

\textbf{MS + IAA}     & BN$^{512}$ & 66.5 & 76.8 & 84.4  &  85.3 & 91.3 & 94.8  & \underline{80.5} & \underline{92.0} & \underline{96.8} & \underline{92.0}  & \underline{98.5} & \underline{99.1}   \\  
\textbf{MS + IAA}     & R$^{512}$  & \underline{68.5} & 78.4 & \underline{86.4}   & \textbf{88.6} & \textbf{93.1} & \textbf{95.8}   & \textbf{82.8} & \textbf{93.2} & \textbf{97.1}  &  \textbf{93.3} & \textbf{98.6} & \textbf{99.2}  \\  \hline  
\end{tabular}
}
\end{table*}

\subsection{Model Robustness}

According to Section \ref{neighbor_correction}, $\alpha$ controls the weight of neighbor correction, and is an important parameter for our method.
Fortunately, $\alpha$ need not to experiment many times to get the best values, since it is a dynamic coefficient relied on the number of samples $n_k$, and computed by our designed neighbor weight function as Eq. (\ref{nwf}):
The larger number of samples within the class is, the smaller value $\alpha$ is, the lower effect of the neighbor correction is applied to the initial covariance matrix.
When $n_k$ in is large, $\alpha$ will close to 0, the neighbor correlation becomes invalid.
Eq. (\ref{nwf}) only depends on an insensitive hyper-parameter. 
(see Section \ref{weight_corr} for more explanations and analysis)

Our method produces the adpative virtual samples for different metric learning losses,
Since it provides more informative samples and extra supervised signals during training,
it can combined with the hard sample minings to boost the performances.
(see Section \ref{hard_mining} for more experimental analysis)


\subsection{Apply to Proxy-based Loss}
In section \ref{iaa}, we have applied our method on the pair-based metric learning loss \cite{schroff2015facenet,wu2017sampling,wang2019multi}, which take pairs of samples to constitute pairwise distances.
We only can apply our method on the proxy-based loss \cite{zhai2018classification,movshovitz2017no,qian2019softtriple}.
Since proxy-based loss does not need large batches and hard mining for boosting, unlike our used pair-based loss relies heavily on them, so the improvements are not so well
(More experimental results and analysis is on Section \ref{more_loss}).

\section{Experiments} 
\label{experiments}

\begin{table}[t]
\centering
\caption{
\textbf{Retrieval performance of our methods compared with the state-of-the-art methods on VehicleID.}
The test sets are splited into three sizes (small, medium, and large). 
}
\label{tab:vehicle}
\resizebox{1. \columnwidth}{!}{
\begin{tabular}{lc|cc|cc|cc}
\hline
 &  & \multicolumn{2}{c|}{Small} & \multicolumn{2}{c|}{Medium} & \multicolumn{2}{c}{Large} \\
Method        & Setting & R@1            & R@5           & R@1            & R@5            & R@1           & R@5           \\ 
\hline \hline

BIER \cite{opitz2017bier}           & G$^{512}$  & 82.6 & 90.6 & 79.3 & 88.3 & 76.0 & 86.4 \\
ABIER \cite{opitz2018deep}         & G$^{512}$  & 86.3 & 92.7 & 83.3 & 88.7 & 81.9 & 88.7 \\

MS \cite{wang2019multi}           & BN$^{512}$ & 91.0 & 96.1 & 89.4 & 94.8 & 86.7 & 93.8 \\
XBM \cite{wang2020cross}          & BN$^{512}$ & 94.6 & 96.9 & 93.4 & 96.0 & 93.0 & 96.1 \\

MIC   \cite{roth2019mic}               & R$^{128}$ & 86.9 & 93.4 &  -   &  -   & 82.0 & 91.0 \\
Divide \cite{sanakoyeu2019divide}      & R$^{128}$ & 87.7 & 92.9 & 85.7 & 90.4 & 82.9 & 90.2 \\  
FastAP \cite{cakir2019deep}                             & R$^{512}$ & 91.9 & 96.8 & 90.6 & 95.9 & 87.5 & 95.1 \\ 
\hline

\textbf{MS + IAA}           & BN$^{512}$              & 95.4 & 97.8 & 93.8 & 96.7 & 92.7 & 96.6  \\
\textbf{MS + IAA}           & R$^{512}$               & \textbf{95.7} & \textbf{97.9} & \textbf{94.3} & \textbf{97.0} & \textbf{93.6} & \textbf{96.9}  \\

\hline
\end{tabular}%
}
\end{table}

\subsection{Datasets}
We evaluate our method on five typical deep metric learning benchmark datasets for image retrieval task.
The training set and testing set of all datasets do not overlap in classes.

\textbf{(1) CUB-200-2011 (CUB)} \cite{WahCUB_200_2011} contains 200 classes of birds with 11,788 images. We split the first 100 classes with 5,864 images for training, and remain the rest of the 100 classes with 5,924 images for testing. 
The images are very similar, some breeds can only be distinguished by minor details.

\textbf{(2) Cars-196 (CARS)}  \cite{krause20133d} contains 196 classes of car models with 16,185 images. We use the standard split with the first 98 classes for training (8,054 images) and the rest of the 98 classes for testing (8,131 images). 

\textbf{(3) Stanford Online Products (SOP)} \cite{oh2016deep} contains 22,634 classes with 120,053 product images downloaded from eBay. We use the standard split as 11,318 classes for training (59,551 images) and other 11,316 classes for testing (60,502 images). 

\textbf{(4) In-shop Clothes Retrieval (InShop)} \cite{liu2016deepfashion} contains 7,982 classes of clothing items with 52,712 images. According to official guidelines, 3,997 classes are for training (25,882 images) and 3,985 classes are for testing (28,760 images). The test set are partitioned to query set and gallery set, where query set contains 14,218 images of 3,985 classes and gallery set contains 12,612 images of 3,985 classes. 

\textbf{(5) PKU VehicleID (VehicleID)}
\cite{liu2016deep} contains 221,736 images of 26,267 vehicles categories captured by surveillance cameras. we use 110,178 images of 13,134 classes for training and 111,585 images of the other classes for testing. We evaluate on the predefined small, medium and large test sets which contain 800, 1,600 and 2,400 classes respectively.

The first two small datasets (CUB, CARS) have \textit{plenty of} samples in each class. The last three large datasets (SOP, InShop, and VehicleID) have \textit{a few} samples.
Hence we apply our neighbor correction on the last three datasets.
More statistical information is shown in Tab. \ref{tab:dataset-info}.

\subsection{Implementation Details}

\subsubsection{Embedding networks}
We implement our method on the single GPU of NVIDIA RTX 2080Ti or 3090.
As previous methods, we use three common backbone networks: GoogLeNet \cite{szegedy2015going}, Inception with batch normalization \cite{ioffe2015batch}, and ResNet50 \cite{he2016deep} pretrained on ImageNet \cite{russakovsky2015imagenet}.
We replace its last layer with a randomly initialized fully-connected layer for metric learning. 
The output embeddings are $L_2$ normalized for computing distance, and the embedding size is $512$.

\subsubsection{Optimization}
The input images are first resized to 256 $\times$ 256, then cropped to 224 $\times$ 224. 
We use random crop and random horizontal flips as common data augment for training, 
and only use the single center crop for test. 
We use Adam \cite{kingma2015adam} with $4 \times 10^{-4}$ weight decay as the optimizer. 
The initial learning rate is $10^{-4}$ and scaled up $10$ times on the output layers for faster
convergence.
We set the training epoch as 100 and scale the learning rate by $0.2$ times during the 40-th and 60-th epoch.
The mini-batches are constructed with the balanced sampler.
The batch size is 128 or 256.

\subsubsection{Hyper-parameters}
We set $M=3$, $\beta=\gamma=0.1$, $K=25$, $\sigma_{m}=\sigma_{cv}=1$, $\tau=40$ as the constant, $\lambda \in [0.6, 0.8]$ based on datasets. 
The estimation of covariance matrix is updated every $4$ epochs, and uses diagonal elements for simplicity.

\begin{table}[t]
\normalsize
\centering
\caption{
\textbf{Performances on informative metrics \cite{musgrave2020metric}.} 
All models use the GoogLeNet as the backbone and 512 as embedding sizes.
}
\label{tab:info_eval}
\resizebox{1. \columnwidth}{!}{
\begin{tabular}{l|cc|cc|cc|cc}
\hline
    & \multicolumn{2}{c|}{CUB} & \multicolumn{2}{c|}{CARS} & \multicolumn{2}{c|}{SOP} & \multicolumn{2}{c}{InShop}   \\   
Method   & MAP & RP & MAP & RP & MAP & RP & MAP & RP \\ \hline \hline

Triplet       & 20.6  & 31.4  & 19.4 & 30.2 & 45.8 & 49.0 & 60.2  & 63.2  \\  
\textbf{Triplet+IAA}        & \textbf{21.5}  & \textbf{32.5}  & \textbf{22.6} & \textbf{33.4}  & \textbf{48.8}  & \textbf{52.0}  & \textbf{63.1}  & \textbf{66.0}  \\  \hline
Contrastive   & 20.8 & 31.9  & 20.3 & 30.5 & 47.7 & 50.8 & 59.9  & 62.8   \\ 
\textbf{Contra+IAA}   & \textbf{22.8}  & \textbf{33.7}  & \textbf{24.6} &  \textbf{34.9} & \textbf{49.7}  & \textbf{52.8} & \textbf{61.9}  & \textbf{64.7}  \\ \hline

MS           & 20.5  & 31.6  & 22.0 & 32.7 & 48.0 & 51.2 & 60.8 & 63.8    \\  
\textbf{MS+IAA}                 & \textbf{22.6}  & \textbf{33.7}  & \textbf{27.6} & \textbf{37.7}  & \textbf{50.5}  & \textbf{53.7} & \textbf{63.3}  & \textbf{66.3}   \\  \hline
\end{tabular}%
}
\end{table}

\begin{table}[t]
\centering
\caption{
\textbf{Comparison with the state-of-the-art deep metric learning methods.} 
Backbones are denoted by abbreviations: BN–InceptionBN, R–ResNet50. 
The superscripts represent embedding sizes. 
}
\resizebox {1. \columnwidth} {!}{
\begin{tabular}{lc|cc|cc|cc}
\hline
  &  & \multicolumn{2}{c|}{CUB} & \multicolumn{2}{c|}{CARS} & \multicolumn{2}{c}{SOP}    \\   
Method & Setting   & MAP & RP & MAP & RP & MAP & RP \\ \hline \hline

Triplet \cite{schroff2015facenet}   & BN$^{512}$      & 23.7 & 34.6 & 23.0 & 33.7 & 43.4 & 46.5    \\
PNCA \cite{movshovitz2017no}       & BN$^{512}$      & 24.2 & 35.1 & 25.4 & 35.6 & 47.2 & 50.1   \\
Margin \cite{wu2017sampling}                             & BN$^{512}$      & 23.1 & 34.0 & 24.2 & 34.8 & 41.8 & 44.9   \\
FastAP \cite{cakir2019deep}                              & BN$^{512}$      & 23.5 & 34.2 & 23.1 & 33.6 & 43.6 & 46.6    \\

MS \cite{wang2019multi}                                 & BN$^{512}$      & 24.7 & 35.4 & 28.1 & 38.1 & 45.8 & 48.8 \\
SNR \cite{yuan2019signal}                               & BN$^{512}$      & 25.8 & 36.6 & 25.0 & 35.2 & 44.5 & 47.4 \\

SR \cite{fu_2021_AAAI}                       & R$^{128}$        & - & 35.1 & - & 37.5 & - & 44.5  \\ 
DIML \cite{zhao2021towards}                  & R$^{128}$        & 23.3 & 34.0 & 23.3 & 33.6 & 42.2 & 45.6   \\ \hline

\textbf{MS + IAA}     & BN$^{512}$  & 25.5 & 36.5 & 29.2  & 39.3  & 54.6 & 57.6   \\  
\textbf{MS + IAA}     & R$^{512}$  & \textbf{26.1} & \textbf{37.1} & \textbf{32.4} & \textbf{42.4} &  \textbf{57.9} & \textbf{60.7}  \\
\hline
\end{tabular}%
}
\label{tab:info_eval2}
\end{table}

\subsubsection{Evaluation protocol}
We follow the same standard evaluation protocol as previous works \cite{proxyanchor,musgrave2020metric}, 
so we evaluate the performances by three retrieval metric. 
\textbf{(1) Recall@K (R@K)}: The fraction of queries that have at least one positive sample in their $K$-nearest neighbors.
\textbf{(2) R-precision (RP)}: For each query, let $R$ be the number of positive samples, RP is defined as the fraction of the number of positive samples in the $R$-nearest neighbors.
\textbf{(3) MAP@R (MAP)}: Similar to mean average precision, but limit the number of nearest neighbors to $R$.
The formal definition is on Supplementary Material.

\subsection{Quantitative Results}

We compare our method with existing sample generation based metric learning methods and all typical state-of-the-art metric learning methods, respectively.

\subsubsection{Comparison with generation-based methods}
We implement baselines and our IAA with GoogLeNet backbone and 512 embedding sizes by following all previous methods of synthetic sample generation.
We also list previous methods with the same backbone and embedding sizes for a fair comparison. 
As presented in Tab. \ref{tab:baslines} and \ref{tab:info_eval}, our method can bring significant performance improvements, such as 3\%-6\% R@1 gains, 1\%-3\% MAP and RP gains on three baselines and five datasets.
Compared with existing generation methods (e.g., DAML, HDML, DVML, Symm, and EE), our method gets obvious performance advantages.
It shows that our method can help to learn discriminative embeddings by leveraging various hard synthetic samples for deep metric learning.

\begin{figure}[t]
\centering
\includegraphics[width=1. \columnwidth]{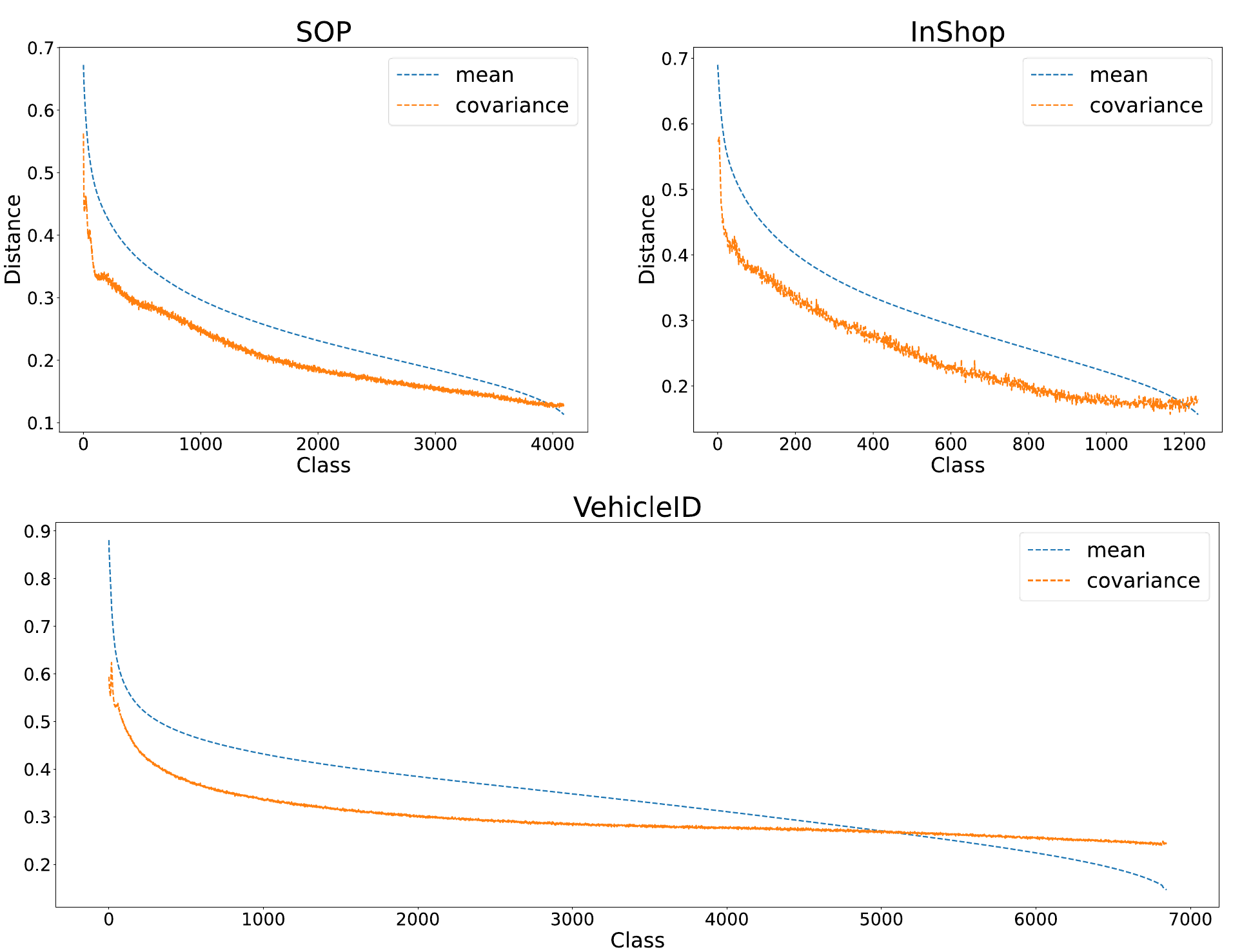}  
\caption{
\textbf{Correlation in other metric learning datasets, SOP, InShop, and VehicleID}. We select the classes whose amount is greater than 5 to compute mean and covariance (last 4233, 1313, and 7022 classes, respectively).
}
\label{other}
\end{figure}

\begin{figure}[t]
\centering
\includegraphics[width=1. \columnwidth]{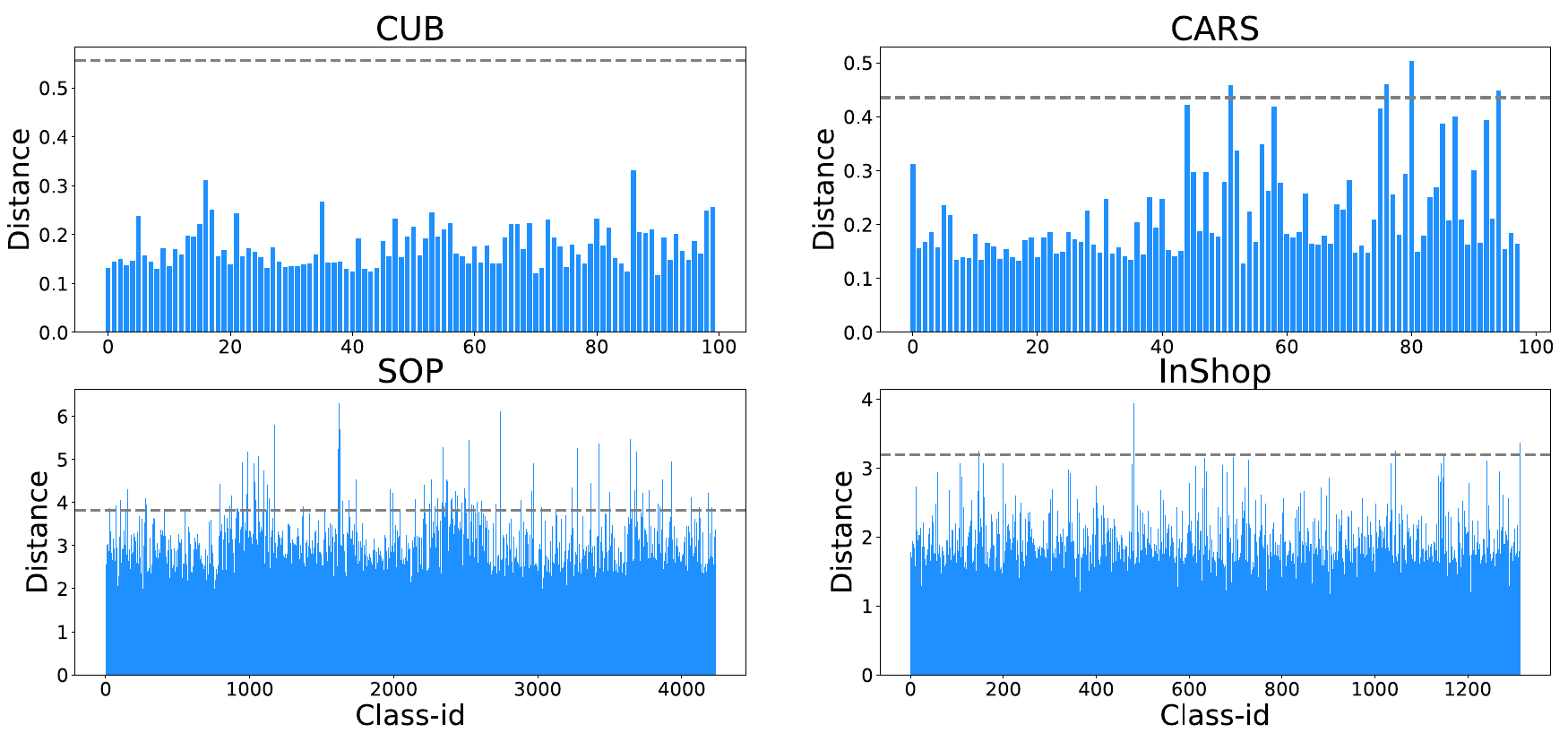}  
\caption{
\textbf{Distances between the global covariance matrix and every class covariance.}
The gray dashed is the value of distance from the global covariance matrix to origin.
For stability, we select classes whose amount is greater than 5 to compute covariance in SOP and InShop.
}
\label{fig:global}
\end{figure}

\subsubsection{Comparison with state-of-the-art methods}
We compare our approach with the state-of-the-art deep metric learning methods.
We write related backbone networks and embedding sizes (generally, larger is better), which affect final performances greatly. 
As shown in Tab. \ref{tab:sota1}, \ref{tab:vehicle}, and \ref{tab:info_eval2}, our method undoubtedly outperforms the state-of-the-art methods on all benchmarks by obvious margins.
We achieve higher performances than loss-designed methods (e.g., PA and Circle)
and boost-like methods  (e.g., DiVA and RaMBO).
Besides, compared with works published within one year (e.g., CGML, SR, and PS), our method still gets preferable performances.
All results demonstrate our method have strong effectiveness and generalization for metric learning.

\subsection{Correlation Analysis}
\label{analysis}
We focus on the authenticity of our correlation discovery, and conduct deep experiments to verify its existence.

\subsubsection{Correlation on more datasets}
\label{more_dataset}
We first conduct the correlation experiments on SOP, InShop, and VehicleID datasets, where the number of class samples is well not distributed.
As shown in Fig. \ref{other}, since SOP and Inshop are few-samples and large datasets, the variance estimation is very inaccurate and there are many classes, so the correlation is not as good as in CUB and CARS, but the correlation does exist.
All similar results show our correlation discovery are indeed consistent despite fields and datasets.

\begin{table}[t]
\centering
\caption{
\textbf{Spearman correlation results between mean and covariance distances by different distance metrics.}.
We use the GoogLeNet backbone trained on each dataset to extract embeddings,
and compute distances by diagonal elements of covariance matrice.
}
\label{tab:spearman2}
\resizebox{1. \columnwidth}{!}{
\begin{tabular}{cc|c|c|c|c|c|c}
\hline
\multicolumn{2}{c|}{\textit{Distance metric}} & \multirow{2}{*}{$p$}  & \multirow{2}{*}{CUB} & \multirow{2}{*}{CARS}  & \multirow{2}{*}{SOP} & \multirow{2}{*}{InShop} & \multirow{2}{*}{VehicleID} \\
mean & covariance &      &   &    &   &    &   \\ \hline  

\multirow{3}{*}{$\left\|\bm{\mu}_i^2 -\bm{\mu}_j^2\right\|_{p}$} & \multirow{3}{*}{ $\left\|\bm{\Sigma}_{i}-\bm{\Sigma}_{j}\right\|_{p}$} & 1  & 0.71 &  0.70   & 0.65  & \textbf{0.71}  & \textbf{0.59} \\
 &  & 2  & \textbf{0.72}  & \textbf{0.72} & \textbf{0.68}  & 0.69 & 0.55 \\
 &  & 3  & 0.69 & 0.71  & 0.56  & 0.66 & 0.53 \\ \hline

\multirow{3}{*}{$\left\|\bm{\mu}_i -\bm{\mu}_j\right\|_{p}$} & \multirow{3}{*}{ $\left\|\bm{\Sigma}_{i}-\bm{\Sigma}_{j}\right\|_{p}$} & 1  & 0.66   & 0.64  & 0.60  & 0.64 & 0.52 \\
 &  & 2 & 0.67  & 0.62   & 0.63  & 0.66  & 0.54 \\
 &  & 3 & 0.65   & 0.60   & 0.61  & 0.65  & 0.51 \\ \hline
 
\multirow{3}{*}{$\left\|\bm{\mu}_i^2 -\bm{\mu}_j^2\right\|_{p}$} & \multirow{3}{*}{ $\left\|\bm{\Sigma}_{i}^{\frac{1}{2}} -\bm{\Sigma}_{j}^{\frac{1}{2}} \right\|_{p}$} & 1  & 0.66   & 0.59   & 0.53  & 0.56 & 0.50 \\
 &  & 2  & 0.68  & 0.64  & 0.55  & 0.58  & 0.48 \\
 &  & 3  & 0.67    & 0.65   & 0.54  & 0.59 & 0.45\\ \hline
 
\multirow{3}{*}{$\left\|\bm{\mu}_{i}-\bm{\mu}_{j}\right\|_{p}$} & \multirow{3}{*}{ $\left\|\bm{\Sigma}_{i}^{\frac{1}{2}} -\bm{\Sigma}_{j}^{\frac{1}{2}} \right\|_{p}$} & 1  & 0.64   & 0.58  & 0.49  & 0.51 & 0.49 \\
 &  & 2 & 0.65  & 0.57   & 0.53  & 0.55  & 0.51 \\
 &  & 3 & 0.64   & 0.56   & 0.50  & 0.52  & 0.43 \\ \hline
 
\end{tabular}%
}
\end{table}

\subsubsection{Distance metric} 
\label{distance_metric}

We show the correlation results by other distance metrics. 
First, we define the distance of different mean by using the \textit{matrix $p$-norm}, with or without square operation.
Similarly, we define the distance of different covariance matrices by using the same the \textit{matrix $p$-norm}, with or without sqrt operation. 
As shown in Tab. \ref{tab:spearman2}, different combinations of distance metrics lead to different degrees of correlation.
We find The first metric is best for measuring the correlation between classes and their variations.
So we select this distance metric to construct the neighbor correction.

\subsubsection{Variety of covariance}
To reveal the diversity of estimated covariance matrix, 
we compute the distance between the global covariance matrix and covariance matrice of each class in Fig. \ref{fig:global}.
We can find most of classes have quite different from global covariance, 
it proves the covariance matrices are diverse as intra-class variations, and not lead to a trivial solution.

\begin{figure}[t]
\centering
\includegraphics[width=1. \columnwidth]{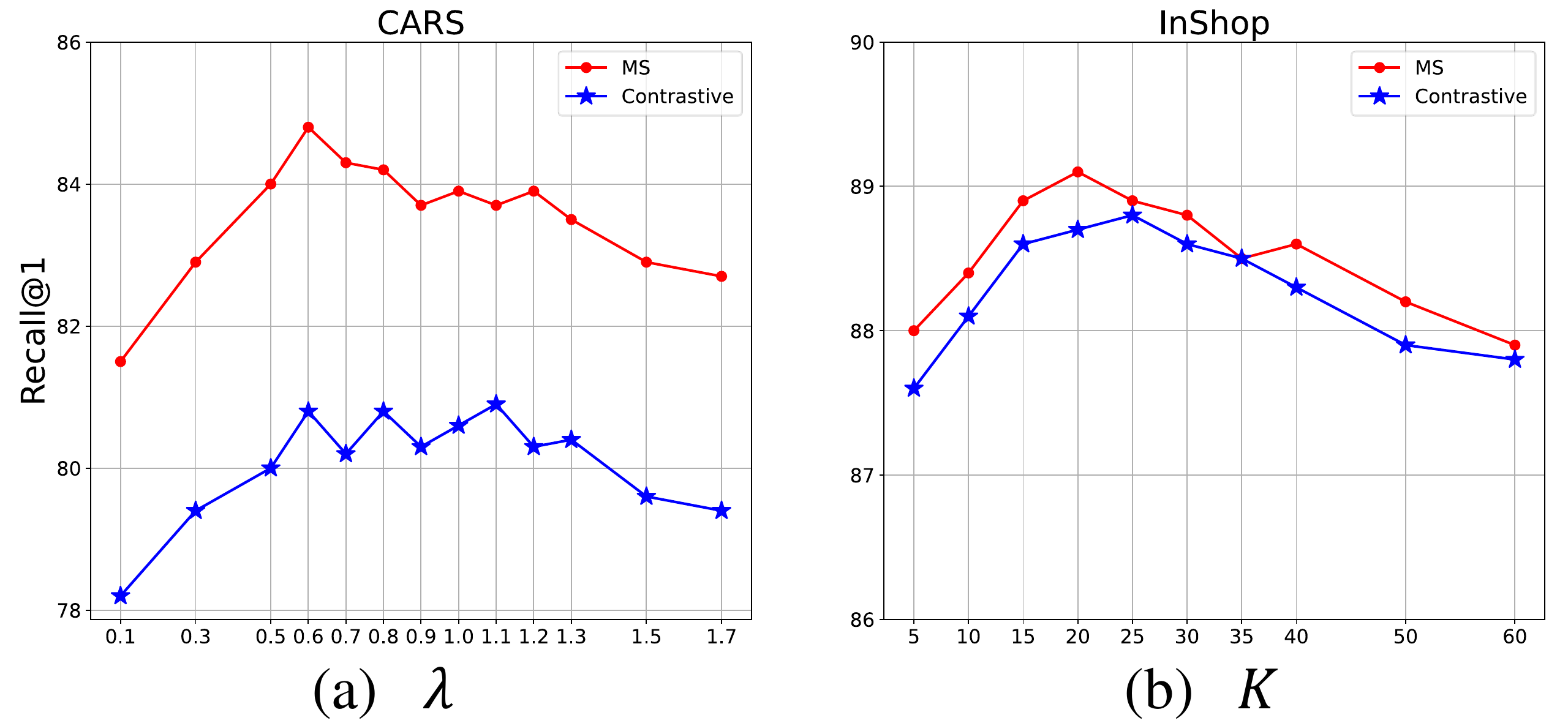}
\caption{
\textbf{Effect of two important hyper-parameters. }
(a) Intra-class variations strength $\lambda$.
(b) The number of neighbor classes $K$.
}
\label{fig:hyper}
\end{figure}

\begin{figure}[t]
\centering
\includegraphics[width=1. \columnwidth]{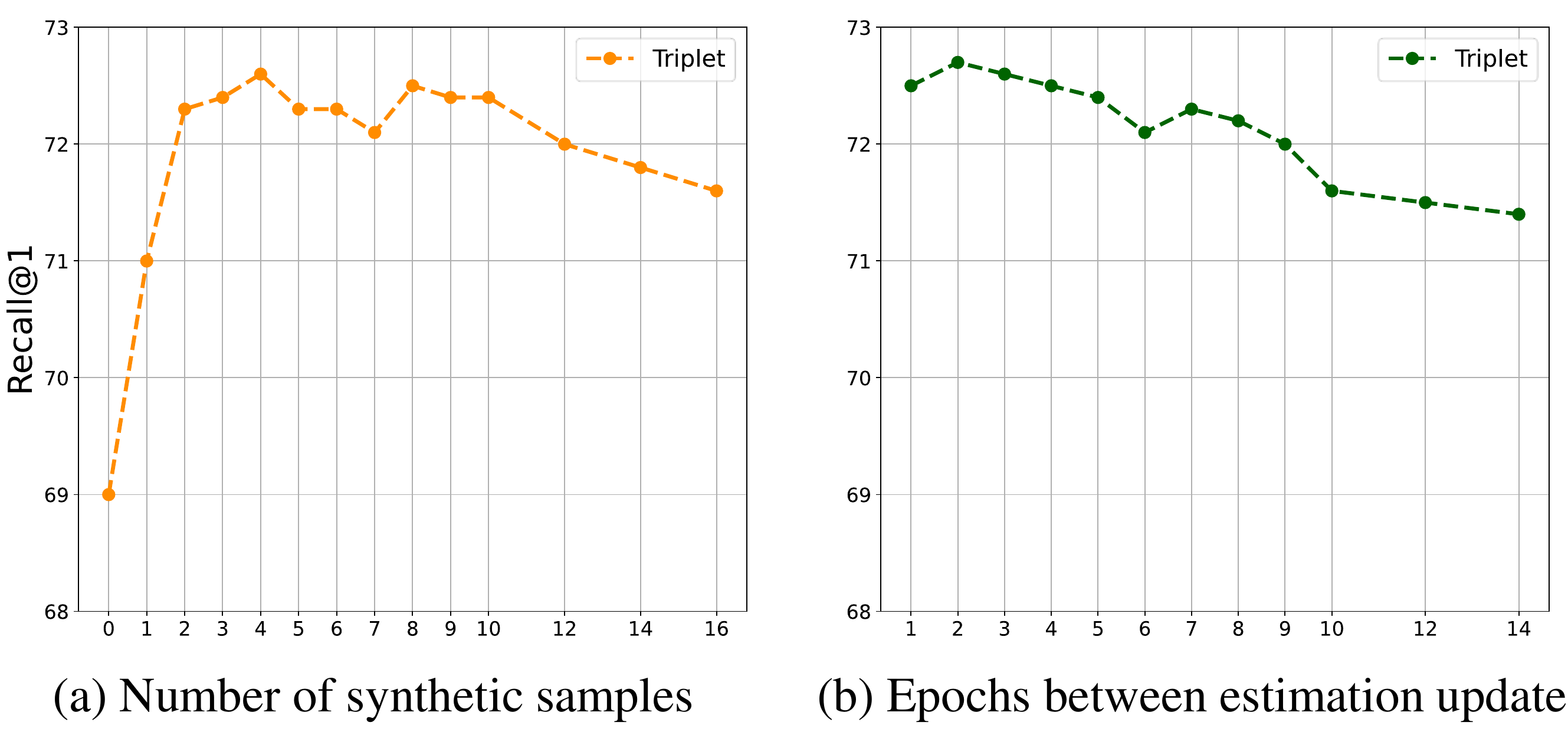}
\caption{\textbf{} 
\textbf{Effect of (a) the number of synthetic samples $M$, (b) the update frequency for estimation} in CARS dataset.
}
\label{fig:syn}
\end{figure}

\subsection{Ablation Study}
\label{ablation}
We provide the ablation study to verify the effectiveness of our method and evaluate the contributions of different parts.
We use the GoogLeNet backbone with 512 embedding sizes and report Recall@1 performance as default.

\begin{table}[t]
\normalsize
\centering
\caption{
\textbf{Effect of various degenerations of the estimated covariance matrix $\bm{\Sigma}_k$.}
\textit{Identity matrix} means replacing $\bm{\Sigma}_k$ by the identity matrix.
\textit{Global matrix} means using the global covariance matrix $\bm{\Sigma}_{ \text { global }}$ to replace $\bm{\Sigma}_k$, 
and \textit{Diagonal matrix} means using the diagonal elements of $\bm{\Sigma}_k$.
 \textit{Correction} means applying the neighbor correction for estimation and only using the diagonal matrix for computing simplicity.
}
\resizebox{0.9 \columnwidth}{!}{
\begin{tabular}{l|cc|cc}
\hline
 & \multicolumn{2}{c|}{w/o Correction} & \multicolumn{2}{c}{w/ Correction }  \\
Setting  & CUB  & CARS  & SOP  & InShop \\ \hline  \hline

  Baseline (Triplet)          & 54.4 & 69.0 & 73.9 & 86.3  \\ \hline 
  
  Identify matrix     & 50.8 & 66.1 & 71.7 & 84.5     \\
  Global  matrix      & 54.9 & 69.8 & 73.1 & 85.8   \\
  Diagonal matrix     & 58.7 & 72.3 & - & -     \\    \hline 
  
  $\alpha=0$  (w/o correction)        & - & -  & 74.2  & 86.9    \\
  $\alpha=1$  (w/o $\bm{\Sigma}_k$)          & - & -  & 76.6  & 88.2   \\ 
  $\gamma=1$  (w/o $\bm{\Sigma}_{ \text { global }}$)          & - & -  & 76.5  & 88.4   \\ 
  $\sigma_{m} = \sigma_{cv} = \infty$                & - &  - &  76.4 & 88.3    \\
  \hline 
 \textbf{IAA}    & \textbf{58.9} & \textbf{72.4} & \textbf{76.9} & \textbf{88.6}  \\ \hline
\end{tabular}%
}
\label{tab:components}
\end{table}

\begin{table}[t]
\centering
\caption{
\textbf{Recall@1 performances on other metric learning loss functions.}
All of them use the GoogLeNet as backbone and 512 as embedding size.
'\checkmark' indicates our method are applied.
}
\label{tab:baslines2}
\resizebox{1. \columnwidth}{!}{
\begin{tabular}{ll|cc|cc}
\hline
&   & \multicolumn{2}{c|}{CUB} & \multicolumn{2}{c}{CARS}  \\
\multicolumn{2}{c|}{More metric learning losses}  &    &   \checkmark  &  & \checkmark \\ \hline  \hline 
 
\multirow{3}{*}{Pair-based} & Triplet+Distance \cite{wu2017sampling}        & 57.7 & \textbf{59.8} & 78.2 & \textbf{80.8}   \\
 & Margin  \cite{wu2017sampling}                                            & 57.5 & \textbf{59.3} & 75.4 & \textbf{77.6}  \\  
 & Circle \cite{sun2020circle}                                              & 60.5 & \textbf{60.9} & 80.8 & \textbf{81.0}  \\
 \hline 
\multirow{3}{*}{Proxy-based}  & Norm-softmax \cite{zhai2018classification}  & 60.1 & \textbf{61.3}  & 80.4 & \textbf{81.8}     \\
 & Proxy-NCA \cite{movshovitz2017no}                                        & 60.5 & \textbf{61.0}  & 81.0 & \textbf{82.1}     \\ 
 & SoftTriple \cite{qian2019softtriple}                                     & 61.0 & \textbf{61.5}  & 82.6 & \textbf{82.8}     \\ 
 \hline 
\end{tabular}
}
\end{table}

\subsubsection{Hyper-parameters}
We show the effect of four important hyper-parameters as shown in Fig. \ref{fig:hyper}.
When one is variable, the other is fixed as default for controlled experiments. 
We find the performances are stable when $\lambda$ is changed in a proper range, 
but can't be too small and large, since it might generate non-informative and invalid synthetic samples. 
When the number of neighbor classes $K$ reduces, the helpful neighbor statistical information will decrease for estimation correction, then related performances drop heavily. 
It proves the importance of our neighbor correction, but selecting too many classes also will introduce noisy signals to hinder correction.

\subsubsection{Number of synthetic samples}
We explore the importance of the number of synthetic samples in Fig. \ref{fig:syn}(a).
Generating $2-4$ samples from each original sample can get superior results,
and the performances do not keep improving by the number of synthetic samples increasing.
It shows our method can provide enough hard and informative samples by the small amount of generations, 
and excessive synthetic samples will distract models from original hard samples.

\begin{figure}[t]
\centering
\includegraphics[width=1. \columnwidth]{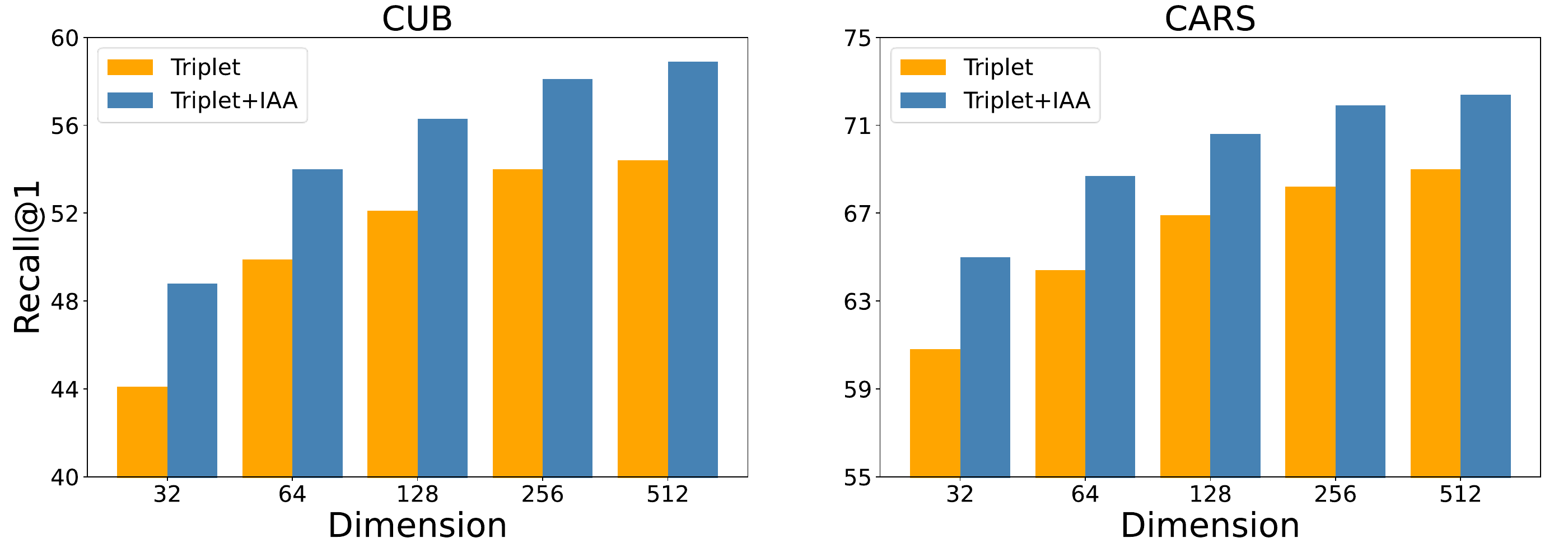}
\caption{
\textbf{Recall@1 comparison on various embedding dimensions.}
 }
\label{fig:emb_size}
\end{figure}

\subsubsection{Update frequency of estimation}
\label{frequency}
Updating the estimated covariance frequently can guarantee the timeliness of distributions during training,
but extra time cost is unacceptable by collecting all embeddings from the datasets.
So we perform on the update frequency in Fig. \ref{fig:syn}(b).
We find performances have no significant drop within a suitable range of epoch interval, 
since the drifting of embeddings is slow in late training \cite{wang2020cross}.

\subsubsection{Loss Function}
\label{more_loss}
Our method can be widely applied to all metric learning losses.
As shown in Tab. \ref{tab:baslines2}, our method also brings performance boost on all benchmarks for other baseline losses,
but the promotion is not so well compared to the results in Tab. \ref{tab:baslines}.
For pair-based losses, we think because the distance-weighted sampling \cite{wu2017sampling} will omit most of informative synthetic samples.
For proxy-based losses, we think because they need not large batch sizes and hard mining, unlike pair-based loss relies heavily on them.

\subsubsection{Embedding dimensions}
In metric learning and various retrieval tasks, the trade-off between speed and accuracy is an important issue, where the embedding dimension is the key factor. 
We test our method with different embedding dimensions, as shown in Fig. \ref{fig:emb_size}. 
Our method signiﬁcantly improves the performance of baselines in all embedding dimensions. 
It indicates that our method constructs a highly efﬁcient embedding space for all the dimensionality.

\begin{figure}[t]
\centering
\includegraphics[width=1. \columnwidth]{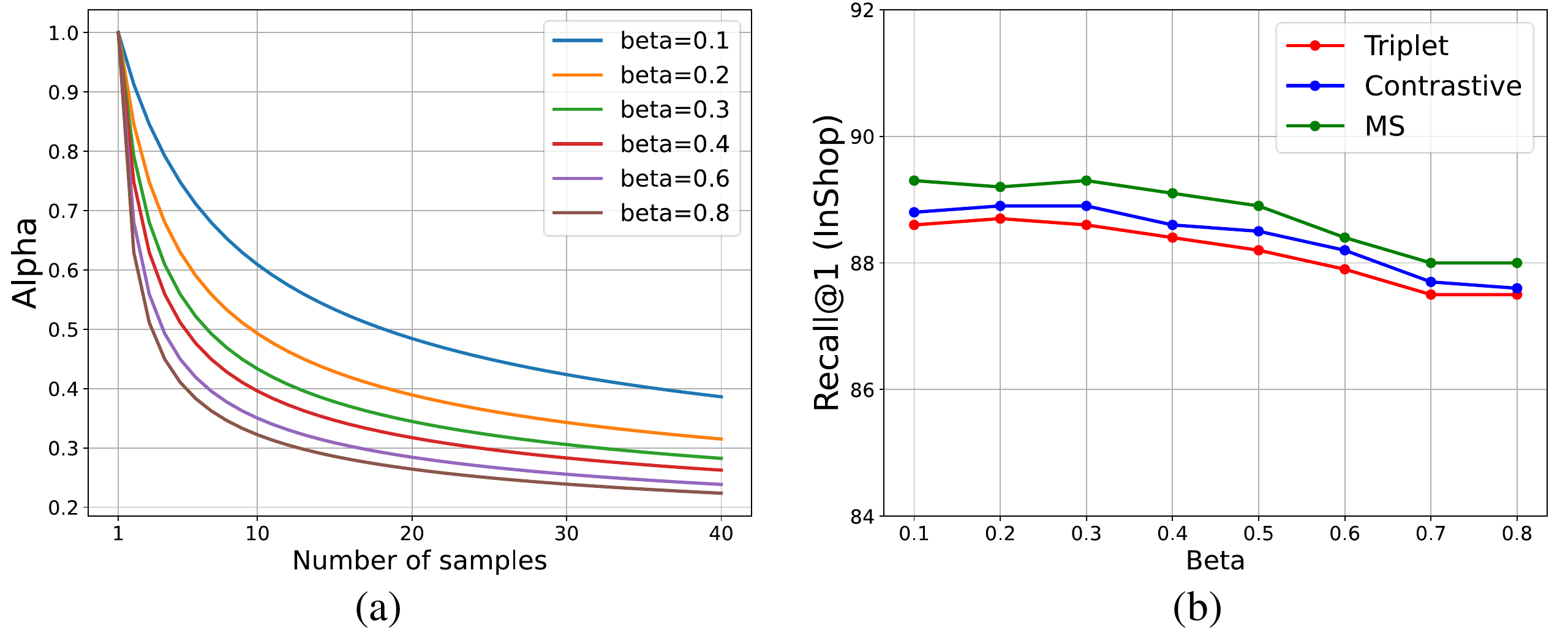}
\caption{
\textbf{The effect of our neighbor weight function}
(a) The $\alpha$ values according to different $\beta$ and the number of samples.
(b) The performances of the parameter $\beta$ of the neighbor weight function.
}
\label{fig:weight}
\end{figure}

\subsubsection{Covariance matrix}
\label{covariance_matrix}
To better verify the effectiveness of estimated covariance matrices, we evaluate them with different degeneration in Tab. \ref{tab:components}.
When we use the identity matrix or single global matrix to replace class-wise $\bm{\Sigma}_k$, the results are terrible, it proves the significance of intra-class variance estimation.
Only using the diagonal matrix can get get the best performance, but the full matrix is the best.
Without our neighbor correction, the defective estimations lead to trivial results.
And the global information is also helpful in preventing overfitting.
Finally, the relative distance weight for neighbor classes improves the performance slightly.

\subsection{Robustness Analysis}
To investigate the availability and robustness of our method, 
we provide more detailed experiments and reasonable analysis.
The experimental settings are the same as Section \ref{ablation}.

\begin{figure}[t]
\centering
\includegraphics[width=1. \columnwidth]{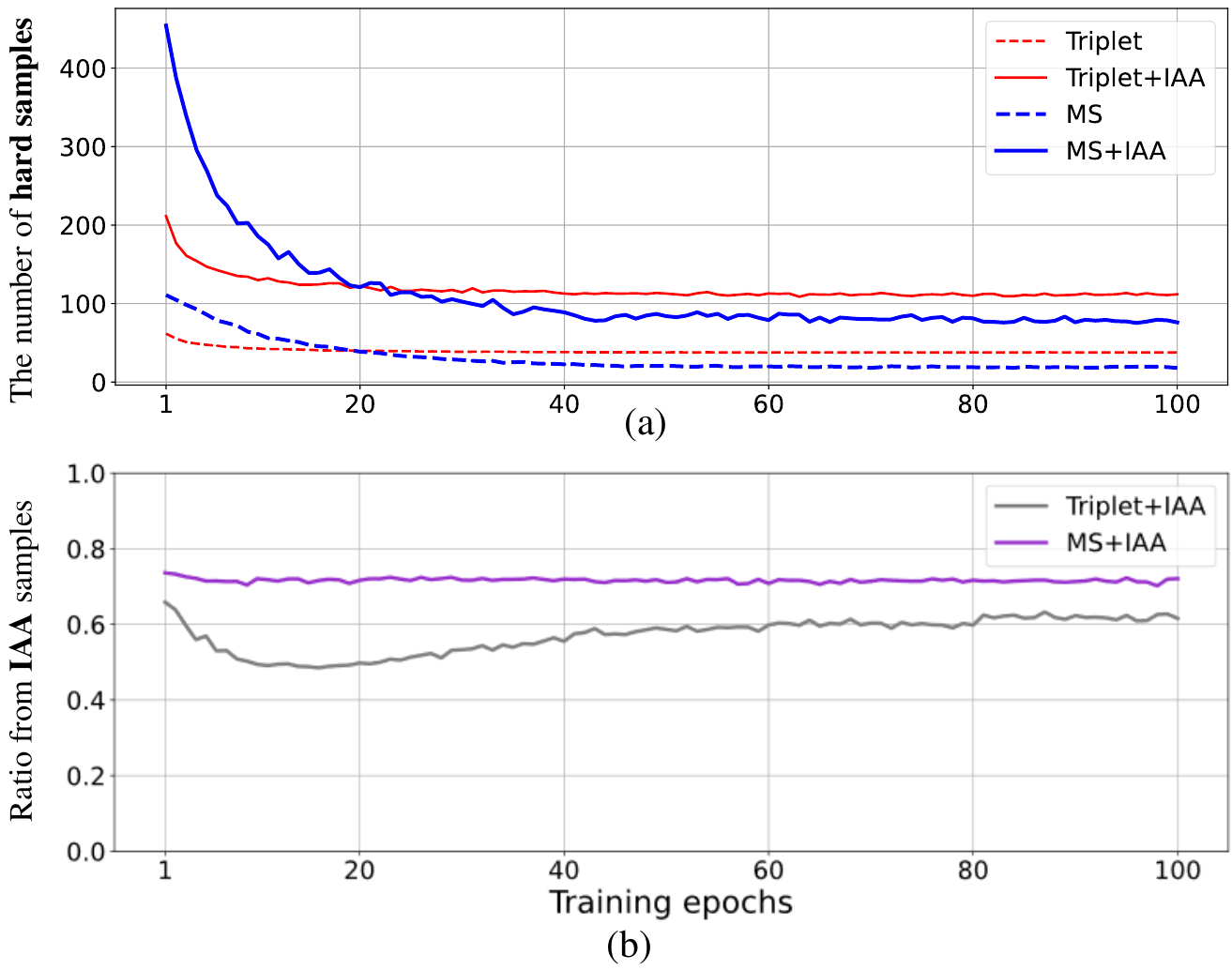}
\caption{\textbf{The effect of hard mining from IAA} 
(a) The number of hard negative samples selected by metric learning losses during training.
(b) Ratio of synthetic sample selected by related hard samples mining during training.
The number of generated samples from each original sample is $M=3$.
}
\label{fig:mining}
\end{figure}

\begin{figure}[t]
\centering
\includegraphics[width=1. \columnwidth]{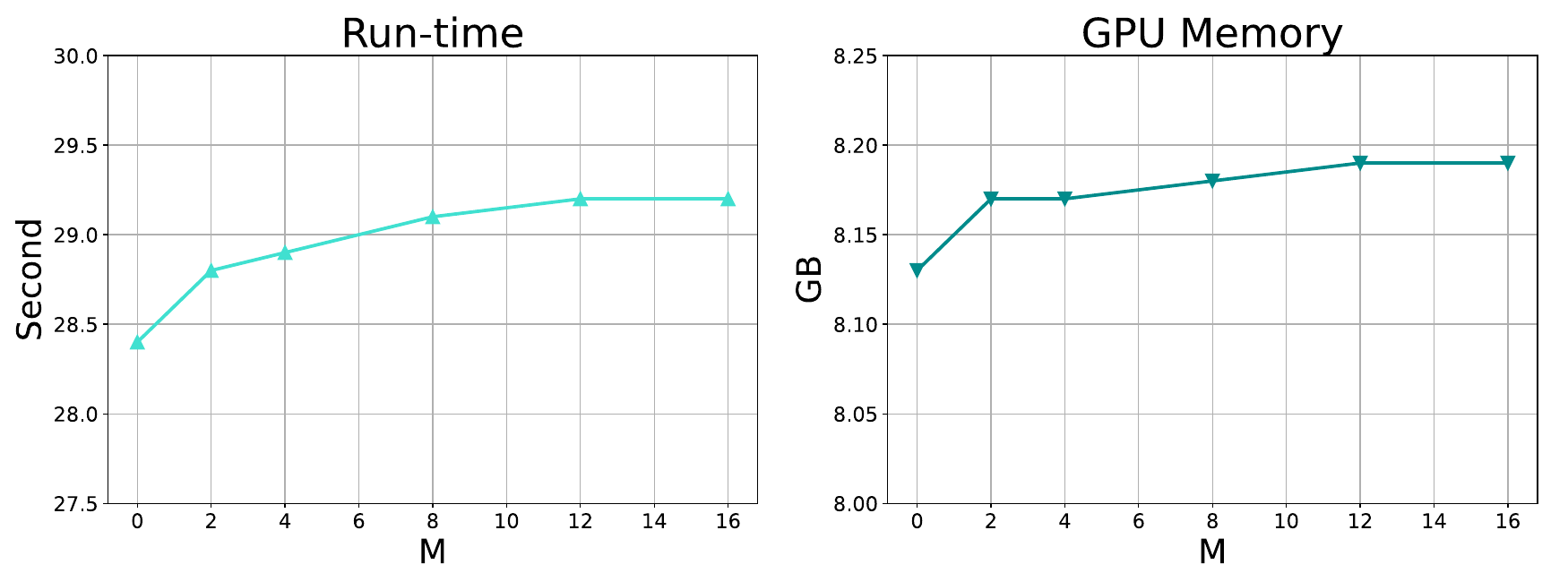}  
\caption{
\textbf{Run-time and GPU-memory cost for sample generation and utilization.}
We train models with Triplet loss and GoogLeNet backbone for 100 iterations. $M$ is the number of generated samples.
}
\label{fig:time_cost}
\end{figure}

\begin{table}[t]
\centering
\caption{
\textbf{Recall@1 comparison with different number of samples $N$, which are selected to estimate the covariance matrix.}
'\checkmark' indicates the neighbor correction are applied.
}
\resizebox{0.8 \columnwidth}{!}{
\begin{tabular}{c|cc|cc}
\hline

 & \multicolumn{2}{c|}{CUB} & \multicolumn{2}{c}{CARS}  \\
Settings  &    &   \checkmark  &  & \checkmark \\ \hline  \hline

Baseline (Triplet)            & \multicolumn{2}{c|}{54.4}  & \multicolumn{2}{c}{69.0}   \\  \hline 

$N=3$           & 55.5  & 56.7  & 70.3 & 71.8 \\
$N=5$           & 55.8 & 57.2  & 70.6  & 71.8  \\ 
$N=7$          & 55.9  & 57.1  & 70.9 & 71.9 \\  \hline 

Use all samples & \textbf{58.9} & 58.7 & \textbf{72.4} & 72.2  \\ \hline
\end{tabular}%
}
\label{tab:nc}
\end{table}

\begin{table}[t]
\centering
\caption{
\textbf{R@1 comparison of different sample generation strategies. }
\textit{dynamic} indicates sampling from $\mathcal{N}\left(\boldsymbol{z}_{i}, \lambda \boldsymbol{\Sigma}_{y_{i}}\right)$, 
and \textit{fixed} indicates sampling from $ \mathcal{N}\left(\boldsymbol{\mu}_{y_i}, \lambda \boldsymbol{\Sigma}_{y_{i}}\right)$.
We train models with Triplet, Contrastive, and MS loss.
}
\resizebox{0.9 \columnwidth}{!}{
\begin{tabular}{l|ccc|ccc}
\hline
 & \multicolumn{3}{c|}{CUB} & \multicolumn{3}{c}{CARS}  \\
Settings  & Trip & Cont &  MS   & Trip & Cont  &  MS \\ \hline

Baseline      & 54.4 & 57.3 & 56.2 & 69.0 & 74.1 & 77.0  \\   

\textit{fixed}           & 54.7 & 55.9 & 57.8  & 70.1 & 74.8 & 79.2  \\
\textit{dynamic}         & \textbf{58.9} & \textbf{61.7} & \textbf{62.1} & \textbf{72.4} & \textbf{80.8} & \textbf{84.2}  \\ 
\hline 
\end{tabular}%
}
\label{tab:gs}
\end{table}

\subsubsection{Weight of correction}
\label{weight_corr}
Although $\alpha$ controls the weight of neighbor correction, it is a dynamic coefficient relied on the number of samples $n_k$, and computed by our designed neighbor weight function as Eq. (\ref{nwf}).
As shown in Fig. \ref{fig:weight},  $\beta$ is the parameter of the neighbor weight function, the smaller it is, the larger the weight is, under the same number of samples.
It proves we need enough neighbor information to correct the defective estimation for optimal results.

\subsubsection{Hard mining for synthetic samples}
\label{hard_mining}
We investigate the hard mining ability by synthetic samples in Fig. \ref{fig:mining}.
We find our method can provides more hard samples for different metric learning losses, and the selection radio of synthetic samples is always high during hard mining process.
Therefore, our generation method can continuously contribute various informative synthetic samples and supervisory signals.

\subsubsection{Time and memory cost}
\label{overhead}
Our method also need additional operations to generate and make use of synthetic samples,
but these computational and memory consumption are almost negligible, since our method has no complex network structure, as shown in Fig. \ref{fig:time_cost}.
Our method just needs \textit{2\%} more run-time cost and \textit{1\%} more GPU-memory cost.

\subsubsection{Availability of neighbor correction}
\label{available_nc}
To further verify the availability of neighbor correction, we apply the neighbor correction on the CUB and CARS
They have adequate number of samples in each class, we just randomly select $N$ samples from every class to estimate the covariance matrices, then use or not the neighbor correction.
Tab. \ref{tab:nc} shows the defective estimation of intra-class variations leads to performance drops for the few-samples class.
Our neighbor correction certainly can alleviate the problem and bring performance promotions by revising the estimation with neighbor statistical information.
However, the approximate method is weaker than the direct estimation from sufficient samples.

\subsubsection{Generation strategy}
\label{generation_strategy}
When we generate synthetic samples from the original sample $\boldsymbol{z}_{i}$, we sampling from the distribution $\mathcal{N}\left(\boldsymbol{z}_{i}, \lambda \boldsymbol{\Sigma}_{y_{i}}\right)$ (dynamic mean) to get diversely semantic intra-class variations and generate synthetic samples.
Besides, we also can sampling from the standard distribution $ \mathcal{N}\left(\boldsymbol{\mu}_{y_i}, \lambda \boldsymbol{\Sigma}_{y_{i}}\right)$ (fixed mean).
Tab. \ref{tab:gs} shows directly sampling from the distribution gets terrible results, even worse than baseline.
Because synthetic samples from this strategy lack of one-to-one correspondence to original samples,
and gradients can't be back propagated from these synthetic samples.



\subsubsection{Similarity distribution}
\begin{figure}[t]
\centering
\includegraphics[width=1. \columnwidth]{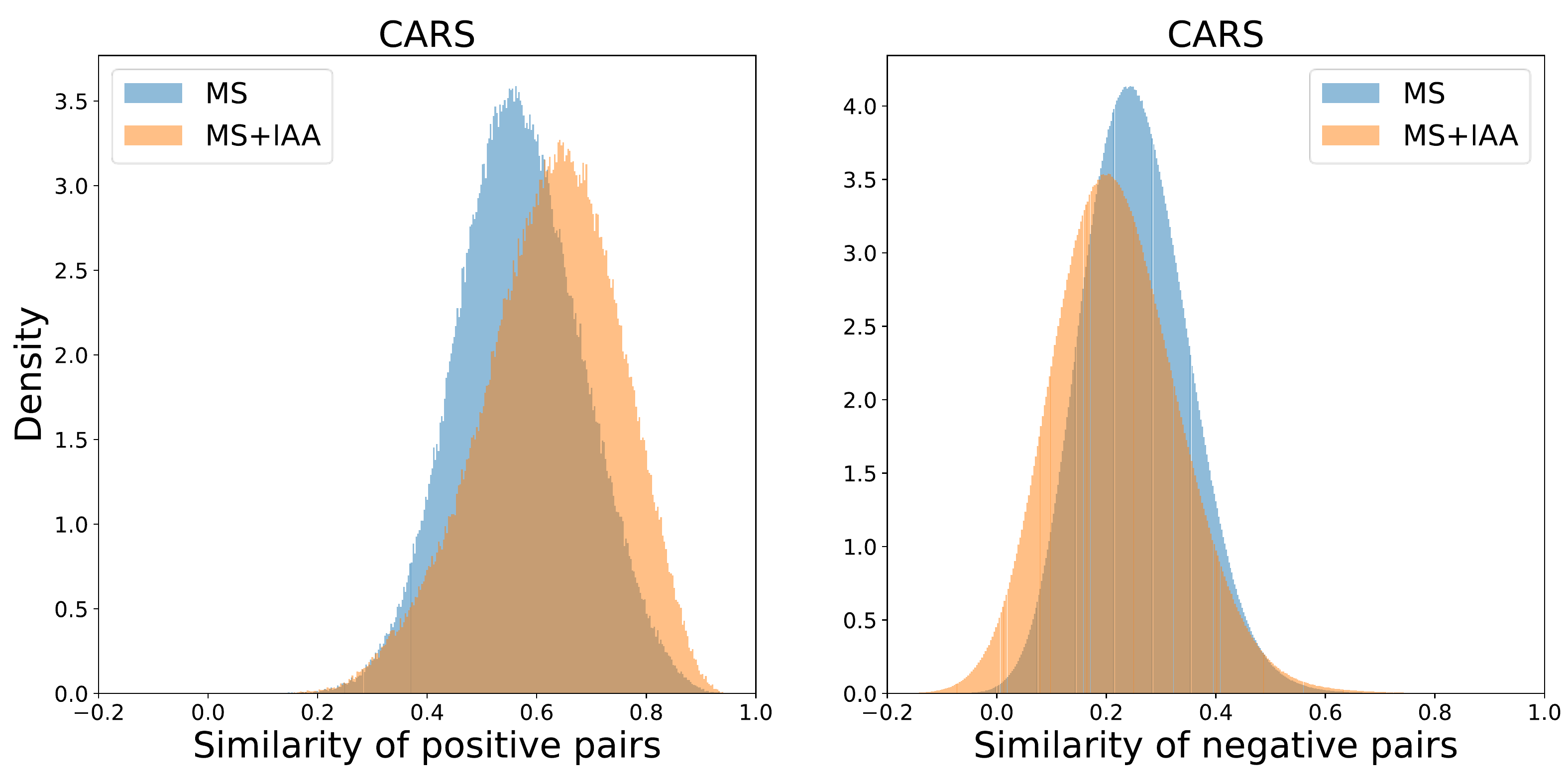}  
\caption{
\textbf{Histogram of cosine similarities of positive pairs and negative pairs with or without our method. }
We take all sample-pairs in the dataset.
}
\label{fig:hist}
\end{figure}

To understand how our method affects the distribution of the similarity, 
Fig. \ref{fig:hist} shows the histograms of similarity of the positive and negative pairs  with or without our method on test set.
It's easy to find that the similarity distribution get improved by our method since negative pairs and positive pairs are separated further.

\begin{figure}[t]
\centering
\includegraphics[width=1. \columnwidth]{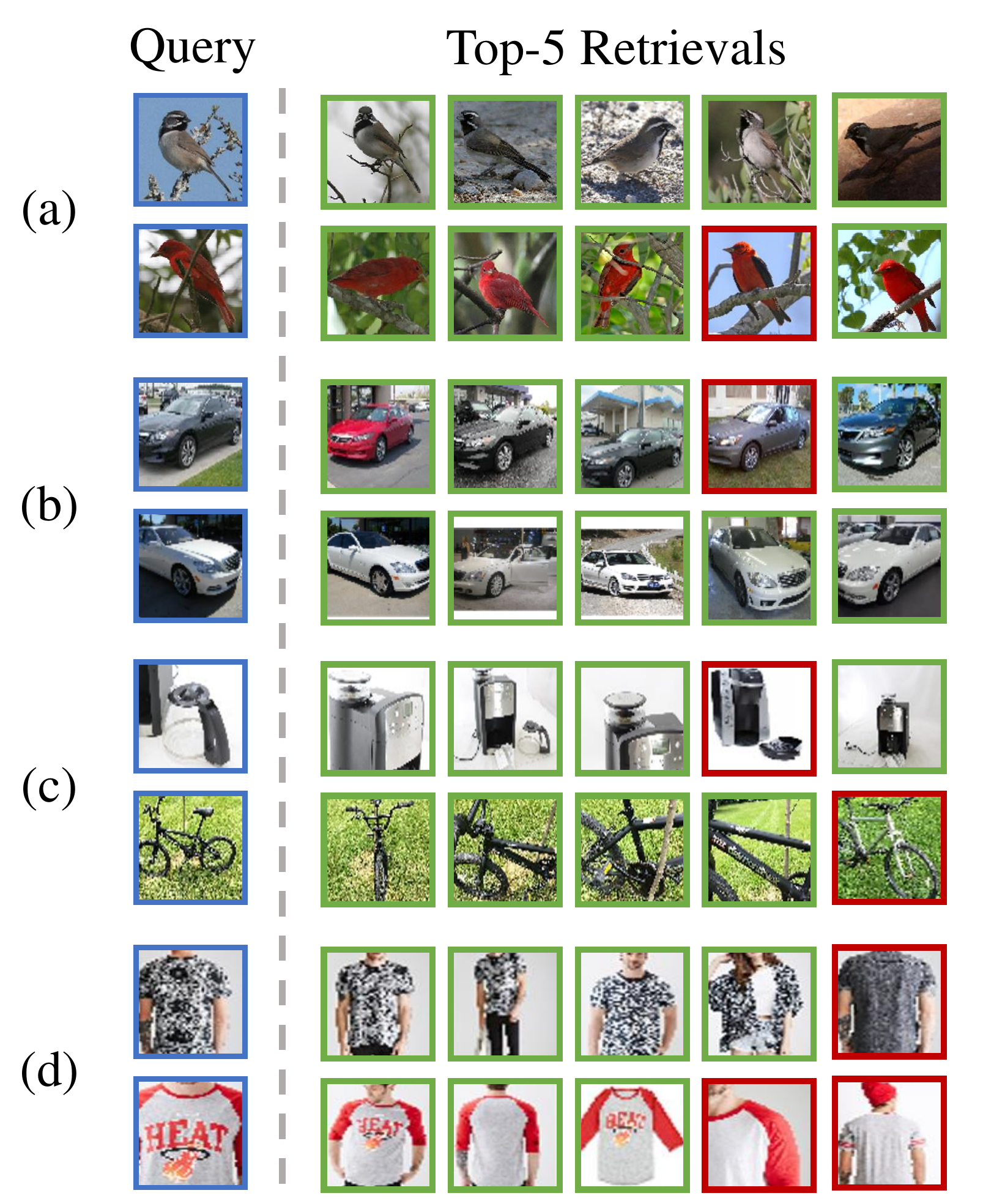}  
\caption{
\textbf{Qualitative retrieval results with our method on MS loss.}
(a), (b), (c), (d) is for four datasets (CUB, CARS, SOP, and InShop) respectively. For each query image (leftmost with blue edge), the top-5 recall results are presented with left-to-right ranking by relative distances. Correct recalls are highlighted with green, while incorrect recalls are red.
}
\label{fig:qualitative}
\end{figure}

\subsection{Visualization}
\label{vis}

\subsubsection{Retrieval results}
We present the qualitative retrieval results of our model on four datasets in Fig. \ref{fig:qualitative}.
Our method can help models to learn more discriminative and robust embeddings, even if intra-class appearance variation is significantly large.
For examples, there are misleading poses (Fig. \ref{fig:qualitative}c), viewpoints (Fig. \ref{fig:qualitative}b), background (Fig. \ref{fig:qualitative}a) , or colors from various semantically similar samples in the same class, but our model achieves great retrieval results qualitatively.

\subsubsection{Embedding space}
\begin{figure}[t]
\centering
\includegraphics[width=1. \columnwidth]{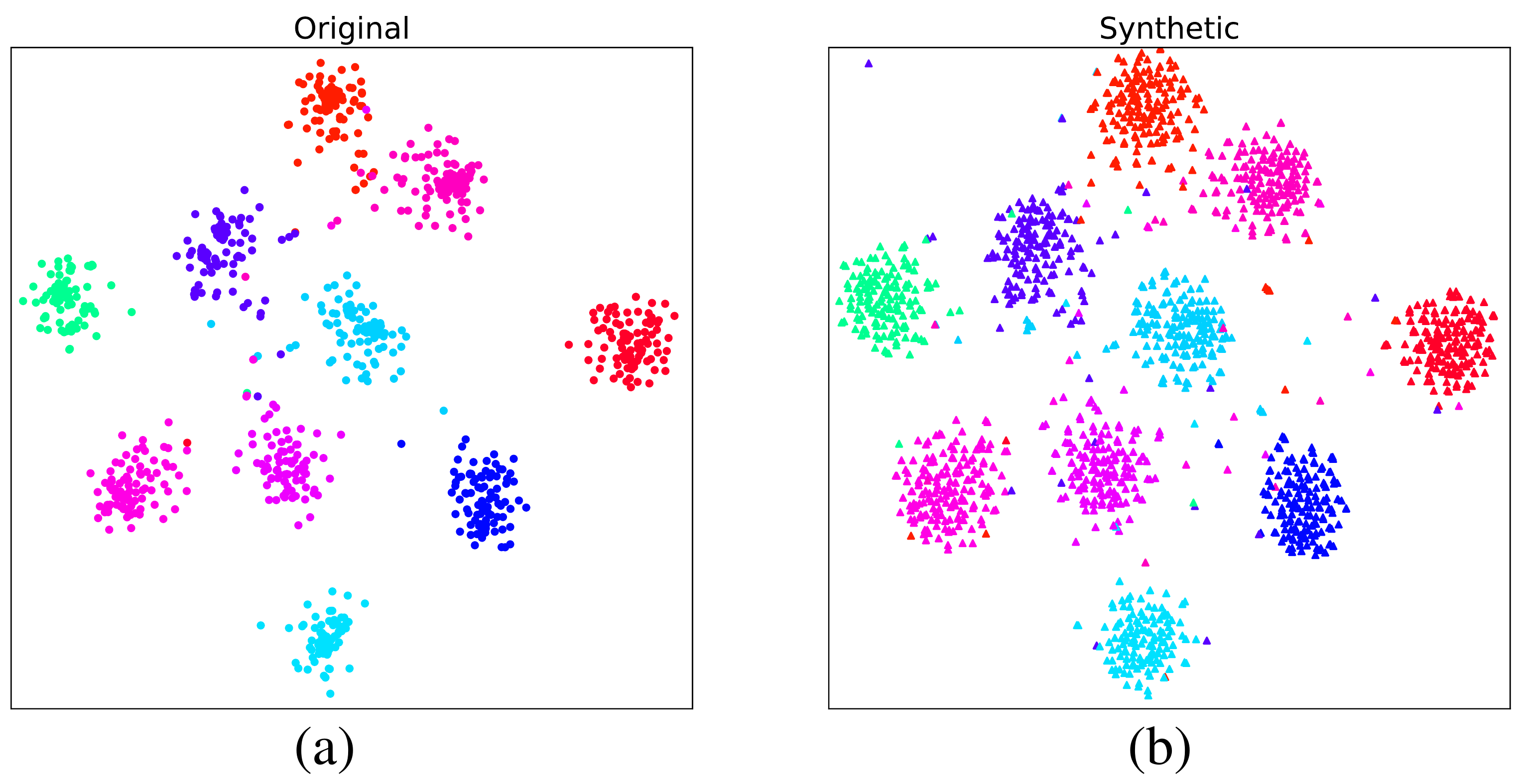}
\caption{
\textbf{T-SNE visualization of the embedding space from 10 randomly selected classes in CARS. }
Different colors represent different classes. (a) Original samples, (b) Synthetic samples. All of them are all from train set.}
\label{fig:tsne2}
\end{figure}

To better qualitatively evaluate the embedding space, we also illustrate the T-SNE \cite{van2014accelerating} visualizations of the embedding space on two datasets. 
As shown in Fig.  \ref{fig:tsne2}, the synthetic samples are generated in-between original samples and overlap most areas of class-clusters, which guarantee the authenticity of synthetic samples. 

\subsubsection{Generation in original domains}
\begin{figure}[t]
\centering
\includegraphics[width=1. \columnwidth]{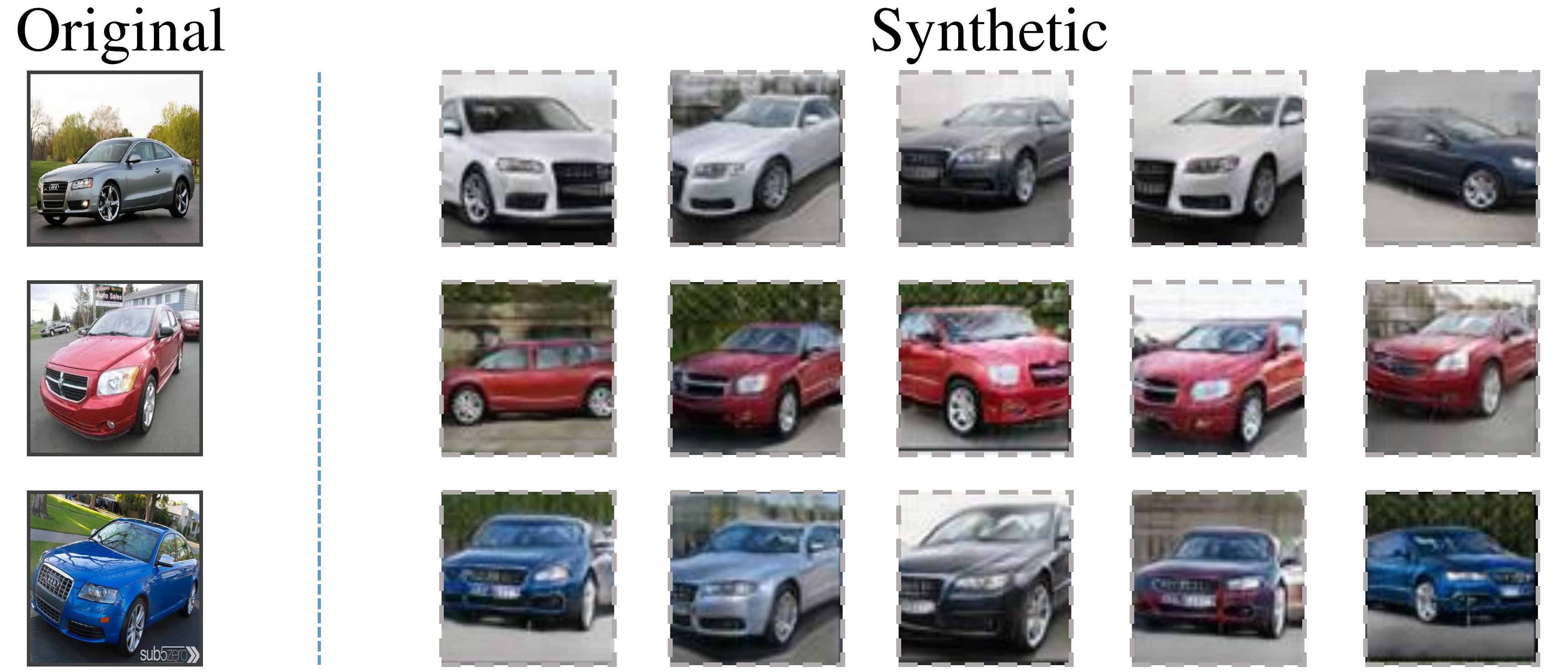}
\caption{ 
\textbf{Visualization of synthetic samples in the pixel space.} 
The first columns are real images in CARS, and rest are related synthetic images by our generation method.
We use the Big-GAN \cite{brock2018large} to map back pixel domain.
}
\label{fig:recons}
\end{figure}

To demonstrate our method generate semantically meaningful samples, we use the utilize the pretrained Big-GAN \cite{brock2018large} to map the synthetic samples back to the pixel space \cite{wang2019idsa} to explicitly show the augmentations of intra-class variations. 
The visualization results in Fig. \ref{fig:recons} show synthetic samples have various of changes of semantic strengths and diversity, e.g., viewpoint, background, and color, which are reliable and robust.

\section{Conclusion} 
\label{conclusion}
In this paper, we present a novel intra-class adaptive augmentation framework for deep metric learning.
We reasonably estimate intra-class variations and generate adaptive virtual samples.
Especially, we discover the similarity correlation between classes and their variations, and propose the neighbor correction to revise inaccurate estimations for the few-samples class.
Our idea of integrating neighbor information provides a new insight for solving defective distribution estimations of insufficient samples, which is a fundamental dilemma for statistical learning.
Moreover, we believe the similarity correlation is a general discovery not only existing in metric learning, and our neighbor correction can be applied to various artificial intelligence tasks.

\bibliographystyle{IEEEtran}
\bibliography{main.bib}

\begin{IEEEbiography}[{\includegraphics[width=1in,height=1.25in,clip,keepaspectratio]{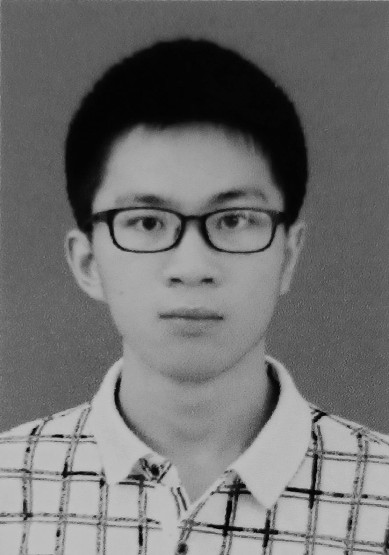}}]{Zheren Fu}
received the B.Sc. degree from University of Science and Technology of China, Hefei, China, in 2020. He is currently working toward the Ph.D. degree with the University of Science and Technology of China, Hefei, China. His research interests mainly cover deep metric learning, cross-modal retrieval, and image-text matching.
\end{IEEEbiography}

\begin{IEEEbiography}[{\includegraphics[width=1in,height=1.25in,clip,keepaspectratio]{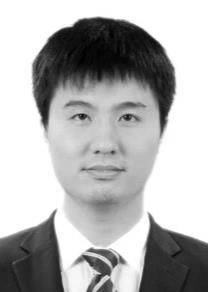}}]{Zhendong Mao} received the Ph.D. degree in computer application technology from the Institute of Computing Technology, Chinese Academy of Sciences, in 2014. He is currently a professor with the School of Cyberspace Science and Technology, University of Science and Technology of China, Hefei, China. He was an assistant professor with the Institute of Information Engineering, Chinese Academy of Sciences, Beijing, from 2014 to 2018. His research interests include computer vision, natural language processing, and cross-modal understanding.
\end{IEEEbiography}

\begin{IEEEbiography}[{\includegraphics[width=1in,height=1.25in,clip,keepaspectratio]{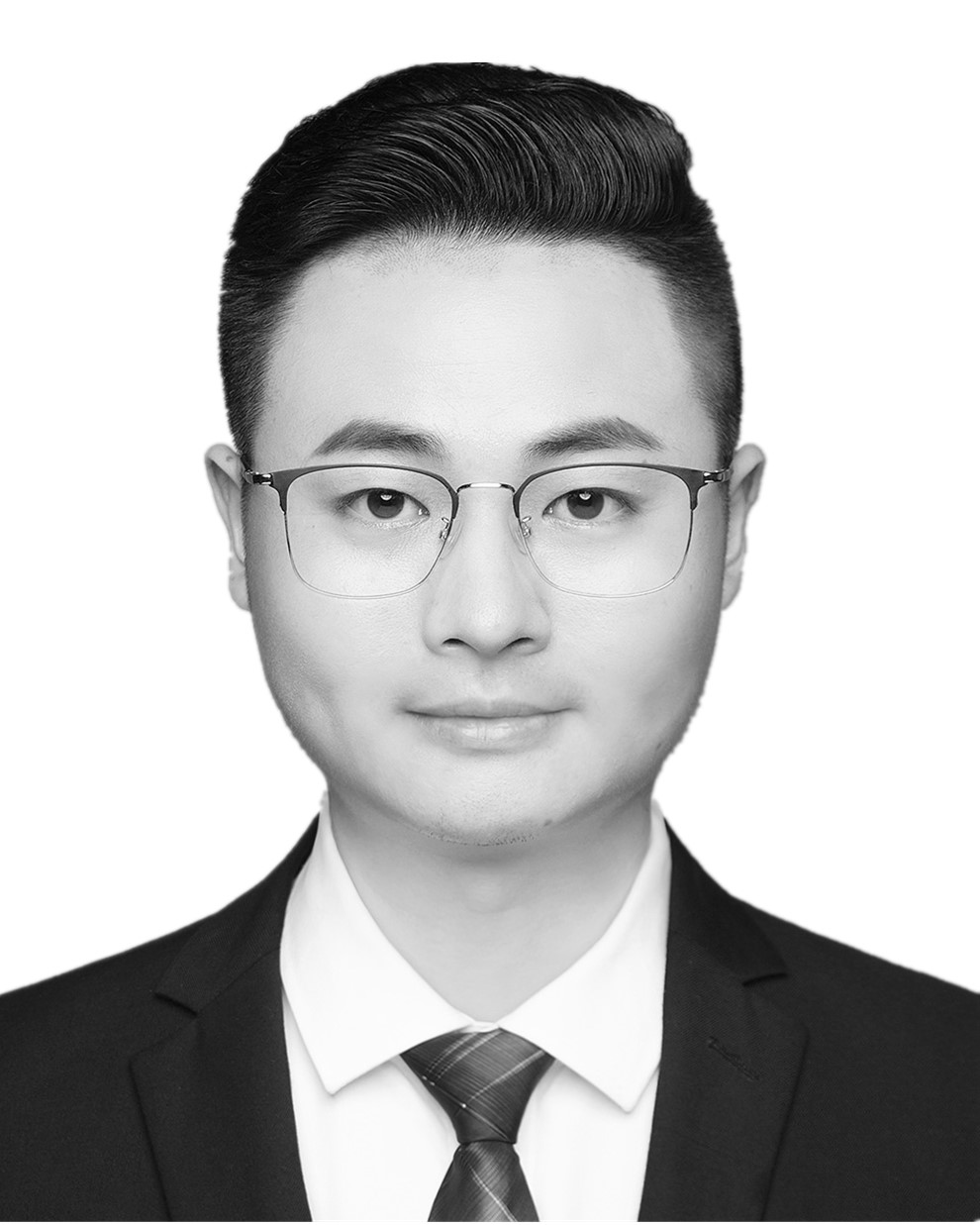}}]{Bo Hu}
received the B.Sc. degree in Computer Science from the University of Science and Technology of China, Hefei, China in 2007, and the Ph.D. degree in Electrical and Computer Engineering from the University of Alberta, Edmonton, AB, Canada in 2013. Currently, he is an associate professor with the School of Information Science and Technology,  University of Science and Technology of China. His research interests include computational social science, recommender systems, data mining and information retrieval.

\end{IEEEbiography}

\begin{IEEEbiography}[{\includegraphics[width=1in,height=1.25in,clip,keepaspectratio]{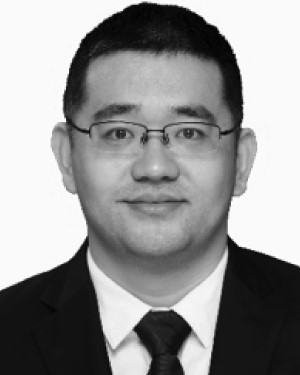}}]{An-An Liu}
received the Ph.D. degree from Tianjin University in 2010. He is currently a Full Professor with the School of Electronic Engineering, Tianjin University, China. His research interests include computer vision and machine learning.
\end{IEEEbiography}

\begin{IEEEbiography}[{\includegraphics[width=1in,height=1.25in,clip,keepaspectratio]{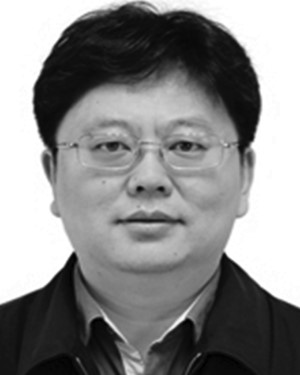}}]{Yongdong Zhang}
(Senior Member, IEEE) received the Ph.D. degree in electronic engineering from Tianjin University, Tianjin, China, in 2002. He is currently a Professor with the School of Information Science and Technology, University of Science and Technology of China, Hefei, China. He has authored more than 100 refereed journal and conference papers. His research interests include multimedia content analysis and understanding, multimedia content security, video encoding, and streaming media technology. He was the recipient of the Best Paper Awards in PCM 2013, ICIMCS 2013, and ICME 2010, and the Best Paper Candidate in ICME 2011. He serves as an Editorial Board Member for the Multimedia Systems Journal and the IEEE TRANSACTIONS ON MULTIMEDIA.
\end{IEEEbiography}

\end{document}